\documentclass[journal]{IEEEtran}

\usepackage{yfonts}
\usepackage[T1]{fontenc}

\usepackage[cmex10]{amsmath}
\usepackage{amssymb} 
\usepackage{mathtools}

\usepackage{graphicx}
\usepackage{subfig}
\usepackage{floatrow}
\usepackage{multirow}
\usepackage{algorithmic}
\usepackage{placeins}
\usepackage{url}
\usepackage{cite}
\usepackage{array}
\usepackage{fixltx2e}
\usepackage{enumitem}
\usepackage{lipsum} 
\usepackage[switch]{lineno} 
\usepackage{xfrac}				
\usepackage{tikz}
\usepackage{pgfplotstable}
\usepackage{pgfplots}
\usepackage[draft]{todonotes}   
\usepackage{comment}
\usepackage{color}
\pdfoutput=1

\def\O{{\mathcal O}} 
\DeclarePairedDelimiterX{\norm}[1]{\lVert}{\rVert}{#1}

\graphicspath{{./fig/}}
\newcommand\redr[1]{{\color{black}#1}} 
\newcommand\redd[1]{{\color{black}#1}} 
\newcommand\red[1]{{\color{black}#1}} 
\newcommand\black[1]{{\color{black}#1}} 

\begin{document}
\title{Accessible Melanoma Detection using Smartphones and Mobile Image Analysis}
\author{Thanh-Toan Do, Tuan Hoang, Victor Pomponiu, Yiren Zhou,  Zhao Chen, Ngai-Man Cheung, Dawn Koh, \\Aaron Tan and 
       Suat-Hoon Tan
	
	\thanks{T.-T. Do, T. Hoang, V. Pomponiu, Y. Zhou,  Z. Chen, N.-M. Cheung and D. Koh are with the Singapore University of Technology and Design (SUTD).  A. Tan and S.-H. Tan are with National Skin Center (NSC), Singapore}

	\thanks{Corresponding author: N.-M. Cheung, SUTD. Email:  ngaiman\_cheung@sutd.edu.sg}
}


\maketitle

\begin{abstract} We investigate the design of an entire mobile imaging system for early detection of melanoma. Different from previous work, we focus on smartphone-captured visible light images.  
Our design addresses two major challenges.  First, images acquired using a smartphone under loosely-controlled environmental conditions may be subject to various distortions, and this makes melanoma detection more difficult.
Second, processing performed on a smartphone is subject to stringent computation and memory constraints.
In our work, we propose a detection system that is optimized to run entirely on the resource-constrained smartphone. Our system intends  to localize the skin lesion by combining a lightweight method for skin detection with a hierarchical segmentation approach using two fast segmentation methods. Moreover, we study an extensive set of image features and propose new \black{numerical features} to characterize a skin lesion. Furthermore, we propose an improved feature selection algorithm to determine a small set of discriminative features used by the final lightweight system. In addition, we study the human-computer interface (HCI) design to understand the usability and acceptance issues of the proposed system. Our extensive evaluation on an image dataset 
provided by National Skin Center - Singapore (117 benign nevi and 67 malignant melanoma)
 confirms the effectiveness of the proposed system for melanoma detection: 89.09\% sensitivity at specificity $\ge$ 90\%. 

\end{abstract}
\begin{IEEEkeywords}
Multimedia-based healthcare, malignant melanoma (MM), mobile image analysis, feature selection, Human-Computer Interface.
\end{IEEEkeywords}
\section{Introduction}

\IEEEPARstart{M}{obile} devices, including smartphones, are being used by billions of people all around the world. 
This creates the opportunity to design a wide variety of mobile image applications, e.g.,
mobile image search~\cite{6719534,6384798,girod:2011}, landmark recognition~\cite{6719514,6725691}, mobile video type classification~\cite{6746214} and 3-D scene video ~\cite{1608117}. 
Among many imaging applications, healthcare applications have drawn a lot of attentions recently.
Several methods ~\cite{6425488,4432626,7506232,baratadermoscopy} have been proposed to support efficient and timely image-related diagnosis.
Apart from normal imaging healthcare applications, mobile imaging healthcare applications have the \black{advantages of being practical, low-cost and accessible}~\cite{hossein:2016,doi:10.1001/jamadermatol.2013.2382,Rosado}.

The work focuses on accessible detection of malignant melanoma (MM) using mobile image analysis.
MM is a type of skin cancer arising from the pigment cells of the epidermis. There are three main types of skin cancers: MM, basal cell carcinoma and squamous cell carcinomas.
Among them, MM is considered most hazardous.
\black{According to an annual~\cite{CAAC}, the American Cancer Society projected 87,110 new cases of melanoma in the United States by the end of 2017, with almost 9,730 estimated deaths.} 
MM may be treated successfully, yet the curability depends on its early detection and removal when the tumor is still relatively small and thin. Therefore, there is a pressing need for tools that can assist early and accurate diagnosis.  

The current practice of  initial melanoma diagnosis is clinical and subjective \cite{Marghoob2009}, relying mainly on the use of naked eye examination. Therefore, the diagnostic accuracy is highly dependent on the trained expertise of dermatologists, which is estimated to be about 85\%~\cite{Sadri2013}. 

\begin{figure}[!t]
\centering
\subfloat[]{
       \includegraphics[width=2cm,height=2cm]{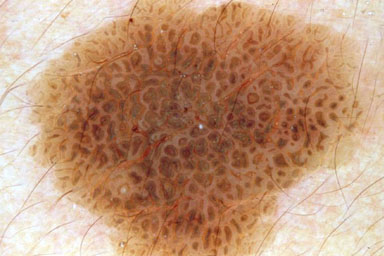}
}
\subfloat[]{
       \includegraphics[width=2cm,height=2cm]{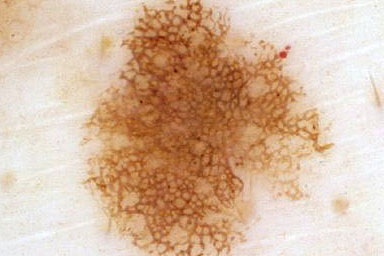}
}
\subfloat[]{
       \includegraphics[width=2cm,height=2cm]{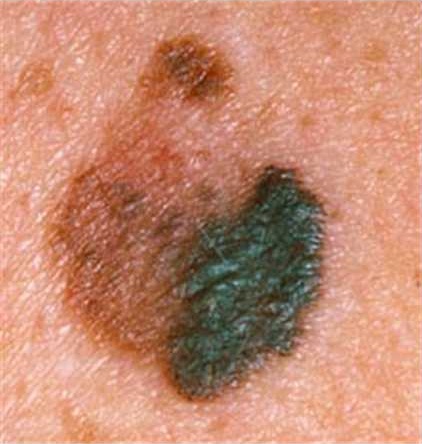}
}
\subfloat[]{
       \includegraphics[width=2cm,height=2cm]{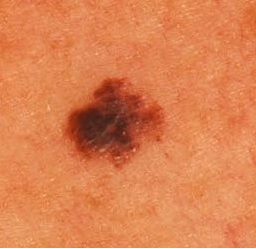} 
}

\vspace{-0.7em}
\caption{Comparison of dermoscopic images (a, b) and visible light images (c, d) of skin lesions.
Dermoscopic images are taken with the aid of liquid medium or non-polarized light source and magnifiers, and they include features under the skin surface (e.g., pigment network, aggregated globules).  On the contrary, visible light images (e.g., taken with smartphones) do not include these features.  This work focuses on the analysis of visible light images.  Source: www.dermoscopy.org, NSC.}
\label{fig:samples1}
\end{figure}
In order to improve the diagnostic identification of MM, dermatologists may use other visual aids such as dermatoscopy~\cite{Sadri2013}.
In clinical evaluation, dermatologists typically apply the ABCDE signs~\cite{ABCDE} (which stands for Asymmetry of the lesion, Border irregularity, Color variation, Diameter, and Evolving), the 7-point checklist~\cite{Argenziano1998}, and the Menzies method~\cite{Menzies1996}, followed by biopsy confirmation of diagnosis.
While diagnosis by dermatologists is accurate, 
a clinic visit
may be less easily accessible and may require the primary physician to make the initial referral.
There is a need for the public to be educated and equipped with a more
accessible method of self-assessment for early diagnosis of melanoma.

Nowadays our industry faces the junction of two rapidly developing markets: healthcare and  emerging mobile computing. The ever-increasing availability of mobile devices equipped with multi-core CPUs and high resolution image sensors have the potential to empower people to become more proactive and engaged in their own healthcare processes~\cite{FDA2013}.

Our contribution is the design of a complete mobile imaging system to detect melanoma.  Our system uses an ordinary smartphone as the platform.  Thus, the proposed system is remarkably accessible.  Our system employs state-of-the-art image analysis algorithms to enable automatic assessment of the 
malignancy of the skin lesion.
Our goal is that  the general public can use our proposed mobile health (mHealth) system to perform preliminary assessment frequently and detect any anomalous skin lesion in their early stage.

As will be further discussed, the proposed system has four major components. The first component is a fast and lightweight segmentation algorithm for skin lesion localization. The second component incorporates new computational features to improve detection accuracy for smartphone-captured skin lesion images.  The third component applies new feature selection tools to
select good features used for the classification. 
The fourth component includes a classifier array and an algorithm to fuse classification results.  
An iterative design approach is used to assess and improve the performance and  utility of the proposed system. We extensively study the system in pre-clinical settings, based on a large number of MM and benign nevi images.

Note that using visible light images captured from smartphones for automatic
melanoma detection is quite new.
Most previous works focused on dermoscopic images that are captured in the well-controlled clinical environments with specialized equipments~\cite{Maglogiannis2009,DBLP:journals/tmi/GansterPRWBK01,DBLP:journals/cmig/CelebiKUIASM07} (including~\cite{Wadhawan2011}, which uses a smartphone for dermoscopic image analysis).
Dermoscopic  images  include features  below  the  skin  surface,  which  cannot  be  captured with  normal  cameras  equipped  in  smartphones (Fig.~\ref{fig:samples1}).
In addition, smartphone-captured images may be subject to various types of distortion (illumination variation, motion blur, defocus aberration).
Therefore, melanoma detection using smartphone-captured visible light images poses some unique challenges, and the problem is not well-understood~\cite{RamlakhanS2011,Doukas2012}.
Furthermore, in order to perform the entire image analysis on the smartphone, we need to design efficient algorithms and system feasible for the strict computation, memory and power constraints of a smartphone.  
To address all these challenges, our works make the following novel contributions \cite{SUTD_EMBC2014}:

\redd{
\begin{itemize}
\item We propose a light-weight skin lesion localization algorithm suitable for the resource-constrained smartphone.  Our localization algorithm comprises skin / non-skin detection, hierarchical segmentation, and combination of Otsu's method and Minimum Spanning Tree (MST) method.

\item We use novel color and border features to quantify the color variation and the border irregularity of skin lesions. We evaluate 116 computational features to quantify the color, border, asymmetry and texture of a skin lesion, including our proposed features that are suitable for  visible light images captured under loosely-controlled lighting conditions.

\item We investigate feature selection to identify a small set of the most discriminative features to be used in the smartphone. Using a small set of discriminative features not only reduces the storage and computation overhead but also improves the classification performance, as low dimensional feature vector is more robust to over-fitting. We focus on the framework using normalized mutual information and propose an improvement that takes into account the feature coordinates.

\item We propose several methods to fuse the classification results of individual category classifiers.

\item We evaluate our system using a dataset from National Skin Center (NSC) of Singapore.  

\item We study the Human Computer Interface (HCI) aspect of the proposed system. 

\end{itemize}
}





The remaining sections of the paper are structured as follows. Section~\ref{sec:overview} reviews melanoma analysis methods. 
Section~\ref{sec:method} presents the details of our proposed system and algorithms. Section~\ref{sec:exp} presents the evaluation and experiments. Finally, Sections~\ref{sec:dis} and~\ref{sec:con} discuss the limitation of the system and conclude the paper by highlighting several future research directions. A preliminary version of this work has been reported~\cite{SUTD_EMBC2014,do_patent}. This paper discusses substantial extension to our previous work: (i) We improve and quantitatively evaluate the proposed segmentation method (Section \ref{subsec:segres}). (ii) We employ and evaluate the Local Binary Pattern descriptor \cite{Guo2010} for the skin lesion analysis (Section \ref{subsec:classificationres}). (iii) We propose new methods for fusing the results of individual category classifiers (Section \ref{subsec:classifier}). The experimental results show that the proposed fusion methods boost the classification accuracy. (iv) We compare the proposed system to the recent melanoma detection methods \cite{mednode,compare_ICIP16} (Section \ref{subsec:comparison}). 
(v) We also study the human-computer interface (HCI) aspect of the proposed system (Section \ref{subsec_HCIres}).

\section{Related Works}
\label{sec:overview}

\begin{figure*}[t]
\vspace{0.2em}
\centering
\includegraphics[scale=0.55]{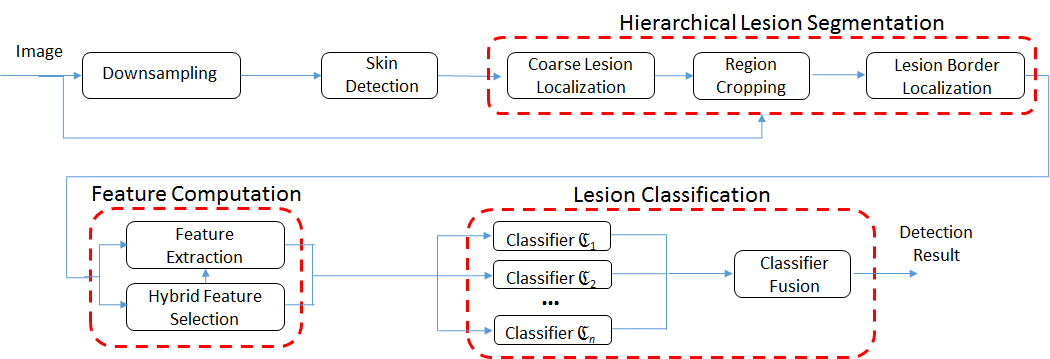}
\vspace{-0.7em}
\caption{The flowchart of the proposed method.}
\label{fig:methodology}
\end{figure*}

Depending on the mechanism used to evaluate the skin lesion, melanoma diagnosis schemes can be classified into the following classes: manual methods~\cite{Menzies1996}, which require the visual inspection by an experienced dermatologist, and automated (computed-aided) schemes~\cite{Garnavi2012, Wadhawan2011, Maglogiannis2009}, 
that perform the assessment without human intervention. A different class, called hybrid schemes, can be identified when dermatologists jointly combine the computer-based result, context knowledge (e.g., skin type, age, gender) and his experience during the final decision \cite{Haider2014}.

In general, an automatic melanoma analysis system can be constructed in four main phases. The first phase is the image acquisition which can be performed through different devices such as dermatoscope, spectroscope, standard digital camera or camera phone. The images acquired by these devices exhibit peculiar features and different qualities, which can significantly change the outcome of the analysis process. The second phase involves skin detection, by removing artifacts (e.g., ruler, hair), and \black{lesion border} localization. The third phase computes a compact set of discriminative features. Finally, the fourth phase classifies the lesions based on the extracted features.
 
There is a plethora of computer-aided systems for segmentation and classification of skin lesions \red{\cite{Garnavi2012,Celebi2013,DBLP:conf/mie/TorreTCG06,Maglogiannis2009,DBLP:journals/tmi/GansterPRWBK01}}. Most of these works investigated for lesion segmentation of a dermoscopic image by using classic image segmentation methods such as histogram thresholding, adaptive thresholding, difference of Gaussian filter, morphological thresholding, wavelet transform, active contours, adaptive snake and random walker algorithm.
However, it is worth noting that these works focused on dermoscopic images which are acquired under controlled clinical conditions by employing a liquid medium (or a non-polarized light source) and magnifiers. This type of images includes features below the skin surface which cannot be captured with standard cameras. 
It is also worth noting that many of these previous works focused on only a certain aspect (e.g., lesion border localization), and they did not provide a complete solution that integrates all the steps. 
In addition, they focused on processing on powerful workstations / servers, where computation and memory are abundant. In the cases of redundant memory and computation capability, recent dermoscopic images-based melanoma detection leverage the power of deep neuron network (DNN) as this could help to achieve very competitive performances \cite{7493284,7493528}.


Recently, several mobile connected dermatoscopic devices have been developed, such as DermLite (3Gen Inc, USA) and HandyScope (FotoFinder Systems, Germany). However, the cost to acquire such an additional device is expensive and they are not accessible to everyone.  Furthermore, trained personnel are required to 
operate dermatoscopic devices.

There are only few systems working on mobile platforms: Lubax (Lubax Inc, CA, USA),~\cite{Gu2014, Abuzaghleh2015,Wadhawan2011,Doukas2012}. However,  many methods merely use the mobile device for capturing, storing and transmission of the skin lesion images to a remote server without performing any computation locally on the mobile device. For example, \cite{Gu2014, Abuzaghleh2015} use a mobile dermatoscope attached to the mobile device to capture dermatoscopic images and send the images to the server for computer assessment.



A few isolated works perform the analysis of smartphone-captured visible-light images directly on the mobile devices. 
In~\cite{RamlakhanS2011}, a mobile-system working for images taken from mobile cameras is presented.
In particular, they presented a preliminary system:  to detect a lesion, they used a very basic thresholding method; to describe the lesion, only standard color feature (such as mean and variance of the color channels, the difference of color through vertical axis) and border features (convexity, compactness) are extracted. 
In~\cite{Doukas2012}, the authors also focused on images taken from mobile cameras. The lesion detection and feature extraction are performed on mobile while the classification can be performed on mobile or cloud. 
However, the authors put more emphasis on the system integration, without mentioning the details of the features used for diagnosis. 
\redr{Recently, the works \cite{8037802,mednode} propose complete systems that segment, extract visual features and classify lesions. In \cite{mednode}, in addition to the automatic extracted color and texture features, additional human annotated information, including lesion's locations, lesion size, number of lesions, etc. is also utilized to  differentiate melanoma from nevocellular naevi. The work \cite{8037802} only uses the automatically extracted features, which are based on the ABCD rule, i.e., Asymmetry, Border, Color, and Diameter features, for the classification process. Different from these works, which only use general features such as mean, variance of different color channels, convexity, solidity, compactness of shape etc., in our work, we propose  novel and robust features specifically for characterization of  lesions. We also propose a novel feature selection method to select a small but very discriminative set of features which not only helps  boost the classification accuracy but also reduce the computation and memory costs. 
Additionally, even though the works~\cite{8037802,mednode} propose complete systems, the efficiency of   these systems running on resource-constrained smartphones has not been reported.  
On the other hand, our method is very efficient on a smartphone as discussed in Section \ref{subsec_HCIres}.} 

\redr{Recently, there are several DNN-based systems that have been proposed for non-dermatoscopic image analysis: lesion segmentation \cite{Jafari2017} and melanoma detection \cite{victor:2016,stanford:2017}.
However, due to the high computation and memory cost of DNN, it is very challenging for such systems to be used on resource-constrained smartphone platforms. 
}

\section{The Method for Early Melanoma Detection}
\label{sec:method}


Fig.~\ref{fig:methodology} depicts our  proposed scheme. 
It is challenging to achieve accurate segmentation of skin lesions from smartphone-captured images under loosely controlled lighting and focal conditions. Instead of using sophisticated segmentation algorithms, which can be computationally expensive, we propose to localize the skin lesion with a combination of skin detection and a fast hierarchical segmentation.
More precisely, we first downsample the skin image, and based on the downsampled version our system generates a coarse model of the lesion by \black{combining two} lightweight segmentation algorithms. Then, to outline the lesion contour, we employ a fine segmentation by using as input the coarse segmentation result. 
From the final segmented region, we extract four feature categories which accurately characterize the lesion color, shape, border and texture. To classify the skin lesion, a classifier is built for each feature category and then the final results is obtained by fusing their results.

\subsection{Lesion segmentation}
\label{sec:segment}

Our segmentation process consists of two main steps. At the first step, a mask of skin regions is generated using the skin detection method. The purpose of the skin detection module is to discard pixels from non-skin regions to simplify the image for the subsequent processing step. At the second step, we extract the lesion by using a hierarchical segmentation method.

\subsubsection{Skin Detection}
\label{subsubsec:skin-det}

The reason for applying a skin detection procedure first is to filter the image from unwanted artifacts, so an exact classification of skin/non-skin regions are not needed as long as we extract the foreground and keep the whole lesion region within. Here we use an approach based on skin color model to detect skin pixels~\cite{Jones1999}. We choose this particular skin model since it is more discriminative, providing 32 skin and non-skin color maps of size $64\times64\times64$ for each skin color. We use the original RGB color image, without any preprocessing, as input to the skin detection model. 

In order to build the skin detection model, we followed the steps given in~\cite{Jones1999}: we first collected a set of  skin/non-skin images to construct our skin detection dataset. Skin images are selected with different skin colors and various lighting conditions for model generalization. The skin color distribution is estimated by a Gaussian mixture model. 
Since the skin mole we want to localize may not have a similar color as the surrounding skin, we use a filling method for all the holes inside the skin region.

\subsubsection{Hierarchical Lesion Segmentation}
\label{subsec:hierarchical-lesion-seg}

Since our objective is to develop a mobile-based diagnosis system, we need a lightweight segmentation method that can achieve high precision under severe computational constraints. Therefore, for the segmentation engine we employ several segmentation methods with \black{low computational requirements}, followed by the use of a criterion to merge the results. It worth noting that the skin lesion images are converted to the grayscale  space for the hierarchical segmentation. 
\paragraph{Coarse Lesion Localization}
\label{subsec:coarse-lesion-loc}

\begin{figure*}[!t]
\centering
\includegraphics[scale=0.6]{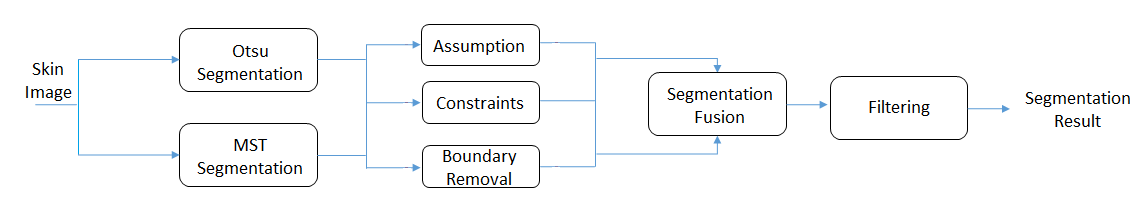}
\vspace{-1em}
\caption{The block diagram of the coarse lesion localization.}
\label{fig:coarse-lesion-loc}
\end{figure*}

After getting the skin region area we downsample the image and perform two segmentation methods and use rules to combine the results of both methods. Here we select Otsu's method~\cite{otsu1975threshold} and Minimum Spanning Tree (MST) method~\cite{felzenszwalb2004efficient} to get the initial segmentation results. Fig.~\ref{fig:coarse-lesion-loc} shows the flowchart of the coarse lesion localization procedure.

\redr{ 
Otsu's method \cite{otsu1975threshold} is a simple and fast thresholding method that automatically calculates a threshold value from image histogram. The threshold value is then used to classify image pixels based on their intensities. However, Otsu thresholding may not detect clear edges on the image, for example, the lesion boundaries.

On the other hand, the Minimum Spanning Tree (MST) method \cite{felzenszwalb2004efficient}, a fast region-based graph-cut method, is sensitive to clear edges. However, MST is not robust  to detect smooth changes of color intensity.}
In our method, the parameters of the MST are optimized such that we could get enough candidate region of interest (ROIs) while avoiding over-segmentation near the skin mole region. As segmentation-based MST algorithm that can run at nearly linear time complexity in the number of pixels, we achieve a low time complexity after running two different segmentation methods, i.e., Otsu's and MST methods.

To filter the segmentation results, we firstly remove all candidate ROIs that are connected to the boundary of skin image. In addition, we assume that the skin mole is located in a region (called the valid region) near the center of the image. This hypothesis is adopted since most of the users focus their camera phone on the object of interest (i.e., the skin mole) when capturing a picture. As a consequence, all the candidate ROIs that have the centroid coordinates outside the valid region are discarded. 
\redd{
Furthermore, we use a constraint to further discard the noisy ROIs. Specifically, for each segmentation method, we identify the ROI by: 
\begin{equation}
\footnotesize
\label{eq:roi_selection}
\underset{i = 1,..., n_R}{\text{argmax}}\left \{  A_i \cdot \left [ 1 - 2 \cdot \sqrt{\left(\mathfrak{C}^{x}_i - \frac{W}{2}\right)^2 +\left(\mathfrak{C}^{y}_i - \frac{H}{2}\right)^2} \right ]^4 \right \}
\end{equation}
where, for the $i^{\text{th}}$ ROI, $A_i$ denotes its area, and $\mathfrak{C}^x_i$ and $\mathfrak{C}^y_i$ are centroid coordinates. $W$ and $H$ are the width and the height of the downsampled image, respectively. $n_R$ represents the total number of ROIs that are located in the valid region. The basic idea is to give central ROIs high weights while penalizing ROIs near to boundary. 
When both $\mathfrak{C}^x_i$ and $\mathfrak{C}^y_i$ are close to the image center, then the value 
$\left( 1 - 2 \cdot \sqrt{\left(\mathfrak{C}^{x}_i - W/2\right)^2 +\left(\mathfrak{C}^{y}_i - H/2\right)^2} \right)^4$ is close to 1.  On the other hand, if the ROI is far from the image center, that value is small. 
The quantity is raised to the fourth power to impose penalty for boundary ROIs.
}
  
By merging the two filtered segmentation results, we expect to get a good segmentation on lesion with either distinct border or diffuse  border. Based on the rules to perform a fusion of different segmentation given in~\cite{DBLP:journals/tmi/GansterPRWBK01}, we take the union of the two results and then find the largest connected region in the union result. The fused segmentation result is post processed by applying a \black{majority filter} to remove the noise. 


\paragraph{Border Localization}
\label{subsec:fine-lesion-loc}

\redd{
Given the coarse segmentation result, in order to correctly localize the border of the lesion, we first crop the corresponding ROI from the original high-resolution image. As  downsampling performed in the ``Coarse Lesion Localization'' step may smear the mole boundary, this mapping is not exact and generates some uncertainty related to mole boundary. 
Thus, another fine-grained segmentation operation is performed to improve the lesion border localization. The segmentation algorithm used for this stage is similar to the one presented in the previous section (i.e., applying Otsu and MST, filtering near boundary segments, and fusing filtered segments), except that we adapt the segmentation parameters to the cropped image characteristics, i.e., a skin image containing a mole which occupies a large part of the image. 
}

\subsection{Feature Descriptors for Describing Lesion}
\label{subsec:feat-des}

Given the lesion image segmented from section~\ref{sec:segment}, we compute features belonging to four categories (color, border, asymmetry and texture) to describe the lesion. The summary of features is given in Table~\ref{tab:feature_cat}. Detail of features are presented in the follows. 

\subsubsection{Lesion Color Feature (LCF) (54 Features)}
Given a skin lesion, we calculate the color features widely used in the literature such as mean, variance of the pixel values of several color channels. The used color channels were \textit{red, green, blue} from the RGB image; \textit{gray scale} from the gray image, and \textit{hue} and \textit{value} from the HSV image. 

To capture more color variation from the skin lesion, we use the information from the histogram of pixel values~\cite{Maglogiannis2009,DBLP:journals/tmi/GansterPRWBK01}. A histogram, with fixed number of bins (i.e. 16 bins), of the pixel values in the lesion is computed and the number of non-zero bins is used as the discriminative feature.  The features generated from the corresponding color channels are called~\textit{num\_red, num\_green, num\_blue, num\_gray, num\_hue} and~\textit{num\_value}. 
\redd{As melanoma samples usually have higher color variation than benign samples \cite{ABCDE}, we expect that the number of non-zero histogram bins of  melanoma samples is higher than that of the benign samples.}\\

\begin{figure}[t]
\centering
\subfloat[Color Triangle]{
       \includegraphics[scale=0.5]{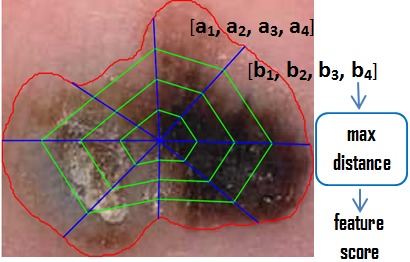}  
       \label{fig:triangle}
}
\subfloat[Border Fitting]{
       \includegraphics[height=2.8cm,width=3.5cm]{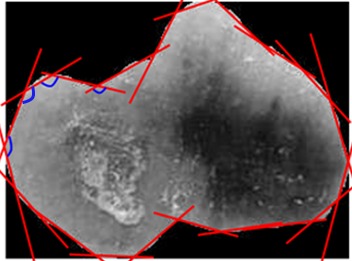}
       \label{fig:border}
}
\vspace{-0.5em}
\caption[]{Novel features to quantify color variation (a): Color Triangle (CT): $SP = 4$ and $PA = 8$ and border irregularity: $nt = 20$ (b) of a skin lesion \cite{SUTD_EMBC2014}}
\label{fig:one_dimension}
\end{figure}

\textbf{Novel feature to quantify color variation:} Generally, there are different color distribution patterns over a MM lesion, whereas the normal pigmented nevi exhibit a more uniform color representation. 
Therefore, another measurement is needed to determine if there is a color variation all over the lesion or the color varies uniformly from the center to the border.
Inspired by the clinical research~\cite{DBLP:journals/cmig/CelebiKUIASM07}, we use a feature called Color Triangle (CT) \cite{SUTD_EMBC2014}. 
\redd{
Specifically, 
given a lesion region (represented by a binary image), we first compute the center of mass of the region. Let $N$ be the number of boundary pixels of the region. Starting from the left most boundary pixel, we divide the region boundary into $PA$ parts in which the number of boundary pixels of each part is equal (i.e. each part has $ N / PA$ boundary pixels). Each triangular part is now represented by two segment lines (from the lesion region center to the boundary). We further equally divide each segment line into $SP$ parts. By connecting corresponding parts on two segment lines, we get $SP$ subparts for each triangular region.


After that, each triangular part is described by a $SP$-component vector, where each component is the mean pixel values of the subpart \cite{SUTD_EMBC2014}. Finally, the maximum Euclidean distance between the vectors is used to quantify the color variation of the skin lesion. As the description, the Color Triangle features are designed to assess  the color spreading  from the lesion center to the lesion boundary. As melanoma mole has non-uniform color spreading \cite{ABCDE}, we expect that when the CT feature is large, the chance of melanoma is high. 
}
In our experiments, the values of $PA$ are empirically chosen as $4, 8, 12$ and $16$. For each value of $PA$, values of $SP$ are chosen as $2, 4$ and $8$. The Color Triangle features are computed for \textit{gray scale}, \textit{red} and \textit{hue} channels of the lesion. Fig.~\ref{fig:one_dimension}(a) illustrates for the proposed Color Triangle feature in which $PA=8$ and $SP=4$. Totally, we extract 54 color features to describe the color variation.  This original feature set is subject to feature selection subsequently to identify a small subset of discriminative features for mobile implementation.

\redd{
Note that we have included mean/variance of the color channel as the candidates in the original feature set. While we expect our new color feature (Color Triangle) to be more discriminative, we have also included mean/variance in the original feature set (before feature selection) for these reasons: 1) mean/variance are conventional features and we have included them as baseline for comparison (during feature selection); 2) mean/variance do not have any parameter to decide, potentially they could be rather general and robust. On the other hand, for histogram or Color Triangle, we need to decide the number of bins / partitions. Nevertheless, as will be discussed, our feature selection results suggest that the generality of mean/variance cannot compensate for their inadequacy in discriminativeness in this application, and mean/variance are not selected in our final feature set.}

\subsubsection{Lesion Border Feature (LBF) (16 Features)}
To describe the irregularity of the border, we compute shape features such as compactness, solidity, convexity, and variance of distances from border points to centroid of lesion \cite{DBLP:journals/cmig/CelebiKUIASM07}.
\redd{Border irregularity is useful for melanoma detection: mole with irregular border has a higher probability  of being melanoma \cite{ABCDE}. 
}

{\bf Novel feature to quantify border irregularity:}
We also propose the Border Fitting feature to quantify the border irregularity \cite{SUTD_EMBC2014}.
The main idea is to approximate the lesion contour by multiple straight lines and then to use the angles between these lines to quantify the irregularity. Regular borders tend to have smooth and consistent changes between consecutive contour fitting lines, compared with the irregular ones. 

We use a linear regression model to determine the contour fitting lines.
We partition the contour (border) points of the lesion into $nt$ segments and $\{x_k, y_k\}$ are the corresponding coordinates for the $i$-th segment. In the linear regression, the predicted coordinate $\hat{y}_k$ can be expressed as $\hat{y}_k = a^i_0+a^i_1 \cdot x_k$, 
where $a^i_0$ and $a^i_1$ are the slope and intercept of the regression line for the $i$-th contour segment.

To quantify the border irregularity, we compute the angles between consecutive regression lines (i.e., regression lines for the $i$-th contour segment and the $(i+1)$-th contour segment). Then, we compute
the average and the variance of the angles, and use these statistics as the border irregularity features. In our experiments, number of lines $nt$ are chosen as
$8, 12, 16, 20, 24$ and $28$. Fig.~\ref{fig:one_dimension}(b) illustrates the Border Fitting feature in which $nt = 20$. Totally, we extract 16 features to describe the border irregularity. 

\redd{
Note that we have included other conventional border features in the original feature set, e.g., convexity, as baseline for comparison (during feature selection), but we expect that they are less discriminative for border irregularity. Our proposed new feature Border Fitting can capture border irregularity well as we directly fit line segments on the border of the mole, and we compute mean / variance of angles between consecutive line segments.  In particular, Border Fitting   can capture notches on the mole border and is useful for melanoma detection \cite{ABCDE}.}

 

\subsubsection{Lesion Asymmetry Feature (LAF) (1 Feature)}
\label{subsubsec:asym}
The lesion asymmetry can also reveal valuable information for the lesion categorization. \redd{According to the analysis in \cite{DBLP:journals/cmig/CelebiKUIASM07}, melanoma samples tend to be asymmetric, while  benign samples tend to be symmetric.} To compute the lesion asymmetry, we use a method similar to  the one introduced in~\cite{DBLP:journals/cmig/CelebiKUIASM07}. The major and minor axes of lesion region, i.e., the first and second principal components, are determined. The lesion is rotated such that the principal axes are coincided with the image axes $x$ and $y$. The object was hypothetically folded to the $x$-axis and the area difference i.e., $A_x$, between the two parts was taken as the amount of asymmetry corresponding to the $x$-axis. We followed the same procedure, for the $y$-axis, to obtain $A_y$. The asymmetric feature is computed as $Asym = \sfrac{(A_x+A_y)}{A}$, 
where $A$ is the lesion area. 

\subsubsection{Lesion Texture Feature (LTF) (45 Features)}
To quantify the texture of the skin lesion, we investigated several feature descriptors such as:
\begin{itemize}
  \item those derived from the gray level co-occurrence matrix (GLCM) and
  \item those based on the local binary patterns (LBP).
\end{itemize}
The GLCM of the entire lesion characterizes the texture by calculating how often pairs of the pixel with specific brightness values and orientation occur in an image. GLCM-based texture description is one of the most well-known and widely used methods in the literature~
\cite{Maglogiannis2009,GLCM}.

In this work, GLCM is constructed by considering each two adjacent pixels in the horizontal direction. The features extracted from GLCM used to describe the lesion are contrast, energy, correlation, and homogeneity. 
To achieve a reasonable estimation of the features, the GLCM should be a dense matrix. Hence, before GLCM calculation, the pixel values are quantized to 32 and 64 levels. It means that we computed 8 texture features from two quantized GLCMs. 

To capture edge map (structure) of the lesion, we employed the Canny edge detector method. The number of edge pixels is counted and normalized by total lesion area and  the resulted number is used as an edge density feature. 

Another widely used texture descriptor that we employed for skin lesion analysis is Local Binary Pattern (LBP)~\cite{Guo2010}.
LBP combines shape and statistical information by a histogram of LBP codes which resemble microstructures in the image. The LBP is a scale invariant measure that describes the local structure in a $3 \times× 3$ pixel block~\cite{journals/pr/OjalaPH96}. The LBP was further adapted to accommodate arbitrary block sizes~\cite{Guo2010}. Two main parameters of LBP are $P$ and $R$, i.e., the number of pixel neighbors $P$ on a circle of radius $R$ of a given center pixel. To generate \red{rotation invariant LBP$_\text{S}$}, $P$-$1$ bitwise shift operations of the circle are performed, and the smallest value is selected. 
In this application, we adopt the LBP framework introduced in~\cite{Guo2010}, since it has a complete mathematical formulation of the LBP operator and it has been extensively tested, offering the best performance. 
Following~\cite{Guo2010}, we extract the sign LBP (LBP$_\text{S}$) from the grayscale channel of the lesion image. The resulted LBP$_\text{S}$ is a 36-dimensional vector. 

\begin{table}[t]
\renewcommand{\arraystretch}{1.1}

\centering
\begin{tabular}{p{1.3cm}p{6.5cm}}
\hline
\hline
Feature Category   & Used features\\
\hline
{}      & Mean, variance of different color spaces (RGB, HSV and grayscale). \\
LCF    
{}        & Number of non-zero bins of the histograms of different color spaces (RGB, HSV and grayscale). \\
          & Color Triangle feature.\\
\hline
{}      & Shape features such as compactness, solidity, convexity \\ LBF           & Variance of distances from lesion border points to the lesion centroid.\\    
{}       & Border Fitting feature.\\
\hline
LAF    & Shape asymmetry.\\
\hline
{}          & Edge density.\\
LTF        & Features computed from the \black{GLCM~\cite{GLCM}} like energy, correlation, contrast, entropy. \\
{}          & Rotation invariant sign and magnitude of \black{LBP \cite{LBP}} of different color spaces (RGB and grayscale).\\
\hline
\hline
\end{tabular}
\vspace{-0.7em}
\caption{Computed features according to their relevant categories and their corresponding descriptions.}
\label{tab:feature_cat}
\end{table}

\subsection{Feature Selection}
\label{subsec:hybrid-feat-sel}

Given the feature set $\mathfrak{F}$ and the class label $L$, we perform feature selection offline to identify a set $\mathfrak{S} \subset  \mathfrak{F}$ 
($\vert \mathfrak{S} \vert \ll \vert  \mathfrak{F} \vert$) 
such that the relevance between $L$ and $\mathfrak{S}$ is maximized. The relevance is usually characterized in terms of Mutual Information (MI)~\cite{Peng05featureselection,Battiti94usingmutual,Estevez:2009:NMI:1657491.1657492}. 
Considering all possible feature subsets requires an exhaustive search which is not practical for a large feature set.

In our method, we use  the well-known feature selection procedure called \red{Normalized Mutual Information Feature Selection (NMIFS)}~\cite{Estevez:2009:NMI:1657491.1657492}. Mutual information is widely employed for the feature selection problem to capture the relevance between variables. In NMIFS, features are selected one by one from 
$\mathfrak{F}$
iteratively.
Initially, the feature that maximizes relevance with target class $L$ is selected as the first feature. 
Given the set of selected feature $\mathfrak{S}=\{f_s\}, s=1,...,|\mathfrak{S}|$, the next feature $f_i \in \mathfrak{F} \setminus \mathfrak{S} $ is chosen such that it maximizes the relevance of $f_i$ with the target class $L$ and minimizes the redundancy between it and the selected features in $\mathfrak{S}$. In other words, $f_i$ is selected by the following condition:

\vspace{-1.5em}
\begin{equation}
\text{argmax}_i\left \{ MI(L,f_i) - \frac{1}{\vert \mathfrak{S} \vert}\sum_{f_s\in \mathfrak{S}}NMI(f_i, f_s) \right \}
\label{eq:selection}
\end{equation}
where $MI(X,Y)$ is the mutual information, which measures the relevance between two random variables \textit{X} and \textit{Y} and is defined as
\vspace{-0.7em}
\begin{equation}
\label{eq:MI_2}
MI(X,Y) = \sum_{x} \sum_{y} {p(x,y)log\frac{p(x,y)}{p(x)p(y)}}
\end{equation}
while $NMI$ is the normalized mutual information and is defined as 
\vspace{-0.7em}
\begin{equation}
NMI(X,Y) = \frac{MI(X,Y)}{\text{min} \{H(X),H(Y)\}} \label{eq:nMI}
\end{equation}
where $H$ is the entropy. From information theory, it is known that $MI(X, Y)\ge 0$; if $X$ or $Y$ is binary variable then $MI(X, Y) \le 1$; and we always have $0 \le NMI(X,Y) \le 1$.\\

\textbf{Novel feature selection criterion \cite{SUTD_EMBC2014}:}
From~(\ref{eq:MI_2}) we observe that the mutual information is a metric that relies on the probability functions, and  it is independent of the coordinate of the features. 
On the other hand, the coordinate of the features may help in the classification context. For example, in a binary classification problem, suppose that the number of samples in each class are equal and there are two features $f_i$ and $f_j$ which perfectly separate the two classes. The feature that has a larger margin between the two classes has a better generalization error. However,  by using mutual information, it can be seen from (\ref{eq:MI_2}) that these features have the same MI value (Fig.~\ref{fig:margin}).
\redd{
 Furthermore, by using the feature coordinates, we can access the local neighborhood structure of data. This also helps the classification \cite{DBLP:journals/pr/WangWZZL09}. 
 }
Therefore, we take into consideration the feature coordinates when selecting the feature.

\begin{figure}[h]
\centering
\includegraphics[scale=0.6]{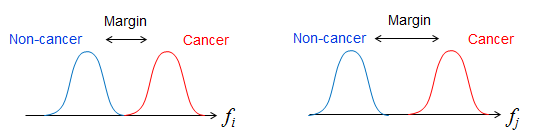}  
\label{fig:margin}
\vspace{-0.7em}
\caption{Two features $f_i$ and $f_j$ that can  perfectly separate the two classes. Feature $f_j$ has a larger margin between the two classes and has a better generalization error. However,  in conventional feature selection, these features have the same MI value following (\ref{eq:MI_2}), as feature coordinates are not considered.}
\label{fig:margin}
\end{figure}

A well-known criterion considering the coordinate of features is Fisher criterion. Nevertheless, there are several issues with the Fisher criterion~\cite{Fukunaga:1990:ISP:92131}.  In particular, if the data in each class does not follow a Gaussian distribution, or  if the mean values of the classes are approximately equal, \red{Fisher criterion fails}.


Inspired by the method ``Average Neighborhood Margin (ANM) maximization''~\cite{DBLP:journals/pr/WangWZZL09} method proposed for face recognition problem, we adapt that method to the feature selection problem and define the quality of feature $f$ as
\vspace{-0.3em}
\begin{equation}
\label{eq:ANM_Q}
\begin{split}
Q(f) &= \sum_{i=1}^{n}
            \norm [\bigg] {\biggl (\sum_{t \in n_i^e}^{} \frac{\norm {f(i) - f(t)}_1}{n_i^e} - 
                                              \sum_{j \in n_i^o}^{} \frac{\norm{ f(i) - f(j)}_1}{n_i^o}\biggr )}_1
\end{split}
\end{equation}
where, for each sample $i$, $n_i^o$ is the set of the most similar samples which are in the same class with $i$ and $n_i^e$ is the set of the most similar samples which are not in the same class with $i$ \cite{SUTD_EMBC2014}. In~(\ref{eq:ANM_Q}) a feature has good discriminative power if we can used it to separate each sample from the samples belonging other classes whilst it is close to samples belonging to the same class. Since (\ref{eq:ANM_Q}) makes use of local information and does not make any assumptions on the distributions of samples, it can overcome the drawbacks of the Fisher test. 

\redd{
In order to combine the advantages of both (\ref{eq:ANM_Q}) (which works on the local neighborhood structure of data) and MI (which works on the distribution of data), we propose the new criterion which combines both (\ref{eq:ANM_Q}) and MI in a single  criterion as follows: 
\begin{equation}
\label{eq:hybrid}
\begin{split}
U(f_i)& = \alpha \cdot Q(f_i) + \\
                 &(1 - \alpha) \cdot \left \{ MI(L,f_i) -
                                  \frac{1}{\vert \mathfrak{S} \vert}  \sum_{f_s \in \mathfrak{S}} NMI(f_i,f_s) \right \}
\end{split}
\end{equation}
where $\alpha \in [0,1]$ is a weight factor that controls the influence of (\ref{eq:ANM_Q}) and MI in the proposed hybrid criterion \cite{SUTD_EMBC2014}. 

}

\subsection{Classifier}
\label{subsec:classifier}

The offline feature selection process is performed individually for each feature category. A compact set of features is identified for each category (color, border, asymmetry, texture: GLCM features and edge density feature).  For a given lesion image, our system computes these selected features for each category and passes them into SVM~\cite{DBLP:journals/tist/ChangL11} classifiers. We train four different SVM classifiers, each corresponding to a single feature category. In this application, we apply a higher penalty for each \red{misclassification} of MM samples than each of benign samples \red{(i.e., the penalty is 1.5 and 1 for MM and benign, respectively)} as we want to achieve high sensitivity while maintaining a reasonable specificity.  

For the LBP features, we use a different classifier. The $k$ nearest neighbor classifier ($k$NN) has been  applied for classification of a majority of LBP descriptors~\cite{Nanni2010} since it is a good candidate when working in a distance representation of objects. Furthermore, since LBP produces a high dimensional feature vector, we will require a large number of samples in order to project the LBP features to higher space where we can use SVM. As a consequence, we adopt $k$NN classifier. 
In general, several metrics have been employed to compute the distance between the LBP histograms such as chi-square distance, L1 distance, and cosine distance. In this study, we adopt the cosine distance metric for the $k$NN classifier since it is more robust to outliers and it has been  widely used in many previous works~\cite{Nanni2010}.

Additionally, it is very helpful to fuse the results from individual classifiers as this combines valuable information gained in the training phase of each feature category. 
Specifically, we propose and experiment these fusion methods:
\begin{enumerate}
\item {\em Non-weighed sum:}
We sum the hard classification results of each classifier (1: cancer or 0: non-cancer). By using the sum rule, we base on the assumption that each feature category contributes equally during the diagnosis decision. Consequently, a skin lesion is judged as cancerous if the sum of the four hard classification values is $\ge 1$ \cite{SUTD_EMBC2014}. 
\item {\em Weighted sum:}
Potentially, individual feature categories have different discriminative power.  Therefore, we fuse their classification results by summing the weighted hard classification values. Particularly, we compute the weights in two different ways: (2a) \textit{sensitivity@50\% specificity}  and (2b) \textit{the Area Under Receive Operating Curve (AUC)}.  We compute these quantities for individual classifiers during the validation phase, and apply these quantities as weights during the testing phase.
For this method, a threshold on the sum of weighted hard classification values is required to decide if a sample is cancerous. We select the threshold that maximizes the accuracy in the validation set.
\item {\em Hierarchical SVM:}
We concatenate the soft classification values of individual classifiers, resulting a new feature vector. The new feature vector is subjected to another SVM for classification. We call this fusion method as hierarchical SVM. This fusion method can uncover the underlying non-linear combination of features in the diagnosis. 
\end{enumerate}

Note that we have two techniques to extract skin lesion texture: GLCM plus edge density, and LBP.
Thus, we experiment these fusion methods for two sets of feature categories: (i) color, border, asymmetry, GLCM plus edge density and (ii) color, border, asymmetry, LBP.

\subsection{Human Computer Interface (HCI) Design}




We also study the HCI aspects of the proposed system.
The proposed system can be regarded as {\em self-diagnosis} applications \cite{forlizzi:chi:2010}.
Self-diagnosis applications refer to a type of accessible healthcare systems that can help users detect illness at the early stages.  Users of self-diagnosis applications are non-patients: people who do not have the explicit awareness of their potential diseases or health problems. For these self-diagnosis applications, it is not clear how to incentivize usage, how to best present the results, and how to sustain habit formation for periodic long-term use. 
To understand these issues, 
we conduct an exploratory case study to investigate the special HCI design challenges of such systems. In particular, we conduct a semi-structured interview with  
16 healthy young people and 13 healthy older adults to investigate the specific HCI challenges of our proposed application.

\section{Evaluation and Results}

\label{sec:exp}

\subsection{Dataset and evaluation protocol}
To evaluate the proposed scheme, we use the dataset provided by the National Skin Center (NSC) of Singapore.
There are totally 184 color images (\textbf{SET1}) consisting of two classes: 117 images of benign nevus and 67 images of MM, acquired by cameras under different resolutions, sizes, and conditions. Due to various acquisition conditions (such as lighting and focus) and  presences of other anatomical features (e.g, eye, eyebrow, nail, etc.) near the skin lesions, many of these images are challenging for the segmentation and classification. 

\red{In regards to} the ground truth for both segmentation and classification tasks, we obtained the diagnosis of the melanoma cases which were determined by histopathological examination or clinical agreement by several expert dermatologists from NSC. Furthermore, an expert was \red{engaged} in the annotation process to help achieve the reliable ground truth ROIs for the skin lesions. 

A smaller dataset called \textbf{SET2}, which is a subset of SET1, i.e. 52 benign images and 29 MM images, is used as the working data for the feature selection phase.

Additionally, we partition SET1 into two subsets. The first subset, called \textbf{SET3},  consists of 30 images, i.e., 18 benign nevi and 12 MM. This subset is used as working data to determine the weight values for \red{the fusion method} 2 proposed in section \ref{subsec:classifier}.
The second subset, called \textbf{TEST SET}, consists of 154 images, i.e., 99 benign nevi and 55 MM. We perform 10-folds cross validation on this TEST SET to obtain the final classification results. We evaluate the classification performance in terms of sensitivity (i.e., Sens = TP/(TP + FN)), specificity (i.e., Spec = FN/(FN + FP)) and balanced accuracy (i.e., Acc = (Sens + Spec)/2).


\subsection{Segmentation Results}
\label{subsec:segres}
The hierarchical segmentation process discussed in Section~\ref{sec:method}
is applied on the SET1 to extract lesion ROIs. 

%
To measure the segmentation results, we used the true detection rate (TDR), which quantifies the rate of pixels correctly classified as lesion. The TDR is computed as follows:
\vspace{-0.3em}
\begin{equation}
\label{eq:TDR}
\begin{split}
\text{TDR(GT, SEG)} = 100 \cdot \frac{\# (\text{GT} \cap \text{SEG})}{\# (\text{GT})}\\
\end{split}
\end{equation}
where SEG denotes the segmentation result and GT denotes the ground truth segmentation.

\begin{table}[!h]
\small

\centering
    \begin{tabular}{*{4}{l}} 
    \hline
    \hline
    Method  & Otsu & MST & Proposed\\ 
    \hline 
    Accuracy(\%) & 72.26 &64.71 & \textbf{80.09}\\
    \hline
    \hline
    \end{tabular}
\vspace{-0.7em}
\caption{Segmentation accuracy on the SET1 of Otsu\cite{otsu1975threshold}, MST\cite{felzenszwalb2004efficient}, and the proposed segmentation method.} 
\label{tab:seg_acc}
\end{table}  
Table \ref{tab:seg_acc} presents segmentation results of Otsu\cite{otsu1975threshold}, MST\cite{felzenszwalb2004efficient}, and the proposed method. It clearly shows that the proposed hierarchical segmentation significantly improves the segmentation results over Otsu and MST methods. 

\begin{figure*}[!t]
\centering
\includegraphics[scale=0.5]{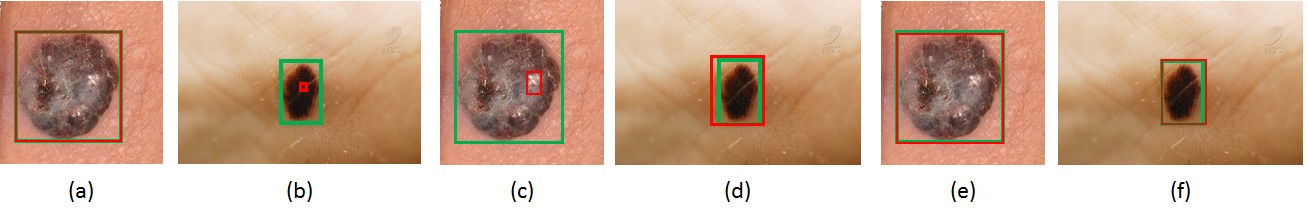}
\vspace{-1em}
\caption{Segmentation evaluation for the Otsu (a), (b), the MST (c), (d), and the proposed (e), (f) methods. The green rectangle represents the ground-truth, the red rectangle denotes our segmentation result.}
\label{fig:seg_results}
\end{figure*}

Fig.~\ref{fig:seg_results} shows the segmentation results for two MM lesions. These are difficult cases where one of the algorithms fails to localize the lesions ROI (for example, Otsu in Fig.~\ref{fig:seg_results}(b), MST in Fig.~\ref{fig:seg_results}(c)). The melanoma lesion shown in Fig.~\ref{fig:seg_results}(a), Fig.~\ref{fig:seg_results}(c) and Fig.~\ref{fig:seg_results}(e) contains several regions with different visual features. This image was well segmented only by the Otsu method (Fig.~\ref{fig:seg_results}(a)), while MST was trapped in one of the  regions of different color intensity (Fig.~\ref{fig:seg_results}(c)). 

For the MM displayed in Fig.~\ref{fig:seg_results}(b), Fig.~\ref{fig:seg_results}(d), and Fig.~\ref{fig:seg_results}(f) the MST method was able to accurately localize its center part (Fig.~\ref{fig:seg_results}(d)) while Otsu method fails entirely (Fig.~\ref{fig:seg_results}(b)). 
However, by using the proposed method we can determine the lesion boundary when one of the segmentation methods fails.  

There are several computation and memory efficient segmentation methods proposed in the literature which may be suitable for mobile-based applications~\cite{tip-level-set,yiren-tcsvt16,gulshan2010geodesic}. Here we give qualitative comparison between the proposed segmentation method with the state-of-the-art level set segmentation method~\cite{tip-level-set} and interactive segmentation method~\cite{gulshan2010geodesic}. 

\redd{
{\bf Comparison with level set segmentation:} Level set segmentation is a fast and accurate segmentation method that has been widely used in many applications.
Here we compare with a state-of-the-art level set segmentation proposed in \cite{tip-level-set} and we use the software by the authors for comparison \cite{code-level-set}.
We found that there are several advantages of our proposed combination of Otsu and MST when comparing with level set segmentation for this application:  1) Level set segmentation would not work well when the boundary of the mole is not clear and the background in the image is complex. As an example from our dataset, Fig.~\ref{fig:143-01}, in this case the boundary of the mole is not clear and the background is not uniform, and the level set segmentation boundary is not accurate compared to our proposed method. 2) There are some parameters needed to be tuned for level set segmentation. As there are different types of images in our dataset (i.e., different background, shape, color of moles), it is difficult to find an optimal set of parameters for our dataset and in general for this application. 3) Our proposed method is time efficient and it is faster than level set. Given the image shown in Fig.~\ref{fig:143-01} for example, it takes 0.4 seconds with our method, and around 4 seconds to run with level set in 200 iterations (both experiments are done on the same desktop and implementation in Matlab).
Note that level set segmentation requires on an initial rectangle provided by the user to crop around the mole region, and its accuracy depends on this initial rectangle.  Thus, it is non-trivial to set up a fair and thorough comparison between level set segmentation and our proposed automatic segmentation. We will investigate this in future work. 

\begin{figure}[!t]
\centering
\subfloat[]{
       \includegraphics[width=0.3\columnwidth]{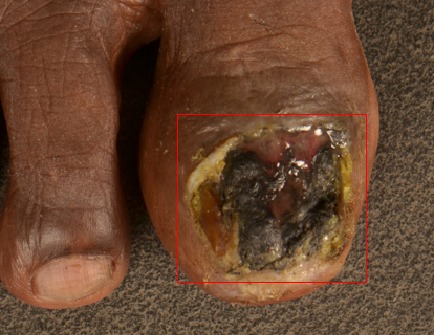}
}
\subfloat[]{
       \includegraphics[width=0.3\columnwidth]{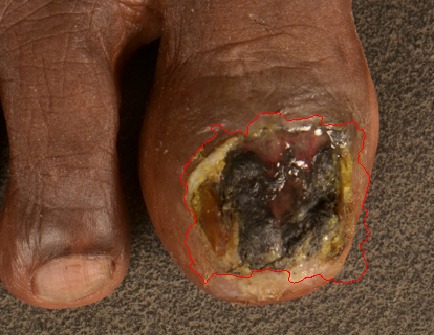}
}
\subfloat[]{
       \includegraphics[width=0.3\columnwidth]{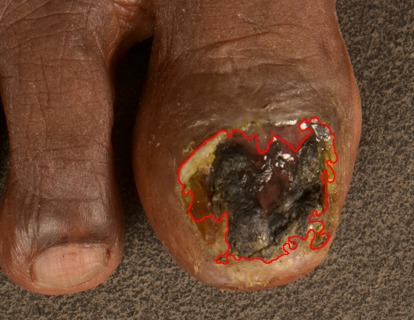}
}
\caption{\redd{Comparison with level set segmentation for an example image with complex backgrounds. (a): cropped 0-level initialization for level set segmentation (provided by the user). (b): level set segmentation result. (c): segmentation result using our proposed method.}}
\vspace{-1em}
\label{fig:143-01}
\end{figure}
}

\redd{
{\bf Comparison with interactive segmentation:}
We have compared with a state-of-the-art interactive method~\cite{gulshan2010geodesic}
for segmentation on our dataset. The code of interactive segmentation is provided by the authors, available from~\cite{code-interactive-seg}. Some comparison results between interactive and our proposed method are shown in Fig.~\ref{fig:graph-cut}. As we can see from the example images, the segmentation results of interactive method depend significantly on the 
 initial strokes  drawn by users. In some cases, even when the initial strokes have only small mistakes, the interactive segmentation results could be quite inaccurate as shown in Fig.~\ref{fig:graph-cut}.  
In order to properly segment the mole, it requires accurate user input, which maybe sometimes difficult for users, especially elders. Therefore, here we apply automatic segmentation method for our application.  
The use of interactive segmentation for this application will be investigated thoroughly in future.
It is worth noting that the accuracy of interactive segmentation depends on the initial strokes drawn by users. Therefore, it is non-trivial to set up a fair and thorough comparison between interactive segmentation~\cite{gulshan2010geodesic} and our proposed automatic segmentation. 
\\
\begin{figure}[!t]
\centering
\subfloat{
       \includegraphics[width=0.24\columnwidth]{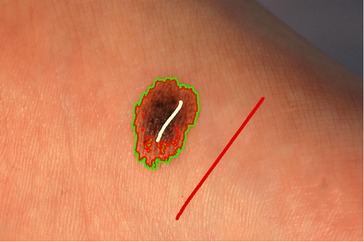}
}
\subfloat{
       \includegraphics[width=0.24\columnwidth]{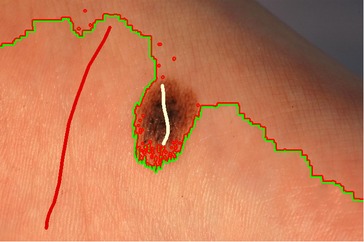}
}
\subfloat{
       \includegraphics[width=0.24\columnwidth]{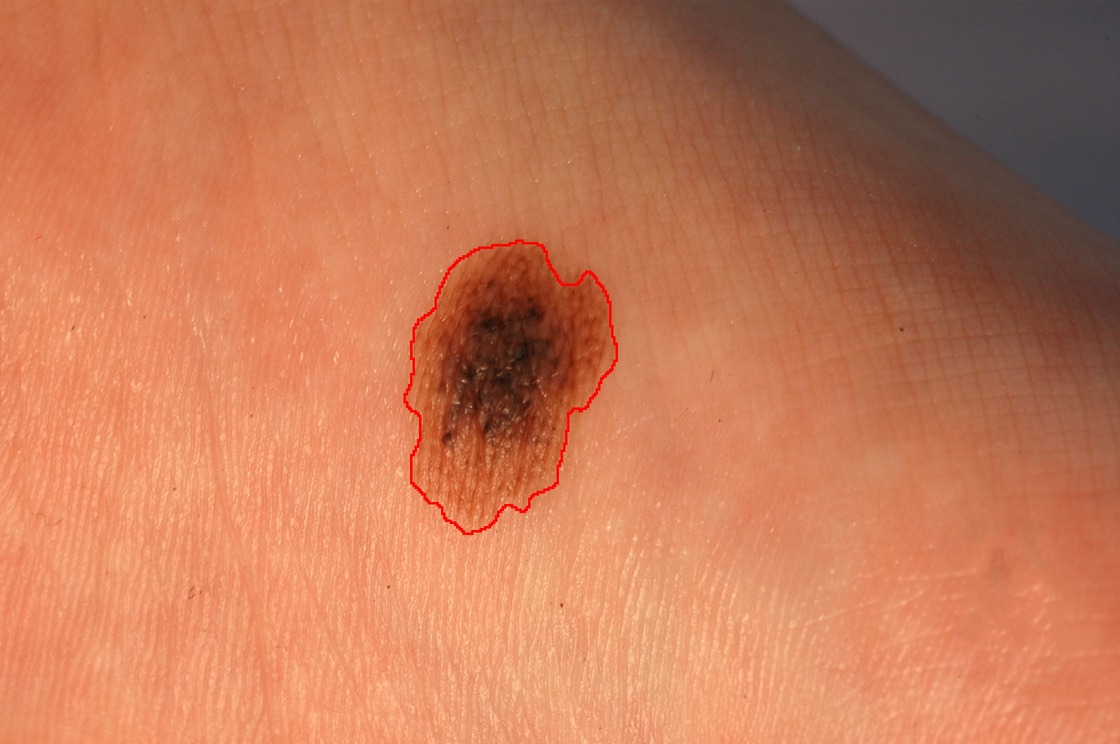}
}

\vspace{-0.7em}
\subfloat{
       \includegraphics[width=0.24\columnwidth]{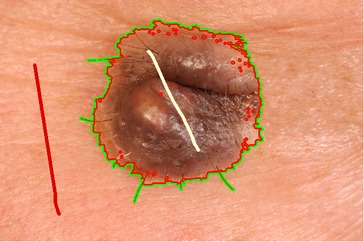}
}
\subfloat{
       \includegraphics[width=0.24\columnwidth]{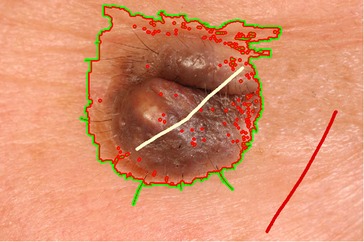}
}
\subfloat{
       \includegraphics[width=0.24\columnwidth]{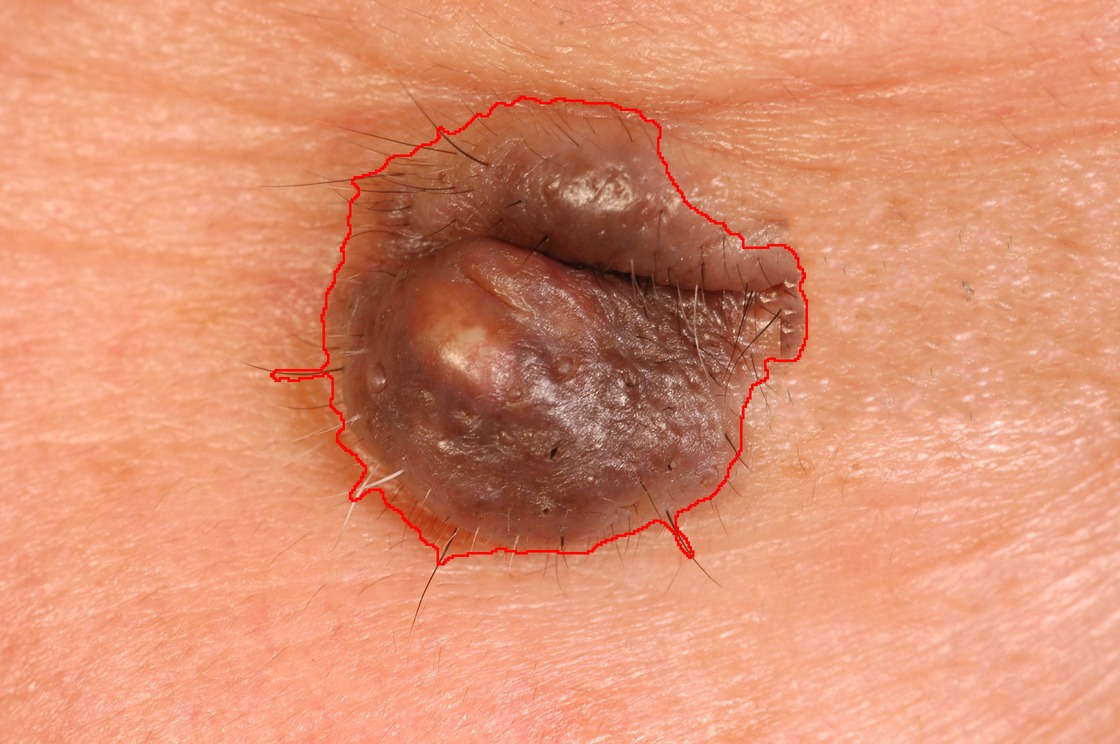}
}

\vspace{-0.7em}
\subfloat{
       \includegraphics[angle=90,origin=c,width=0.24\columnwidth]{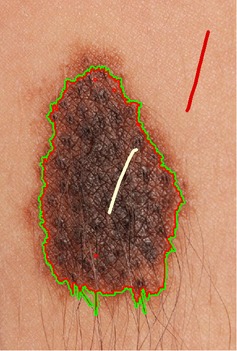}
}
\subfloat{
       \includegraphics[angle=90,origin=c,width=0.24\columnwidth]{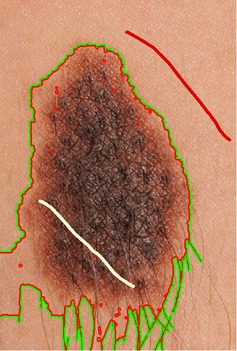}
}
\subfloat{
       \includegraphics[angle=90,origin=c,width=0.24\columnwidth]{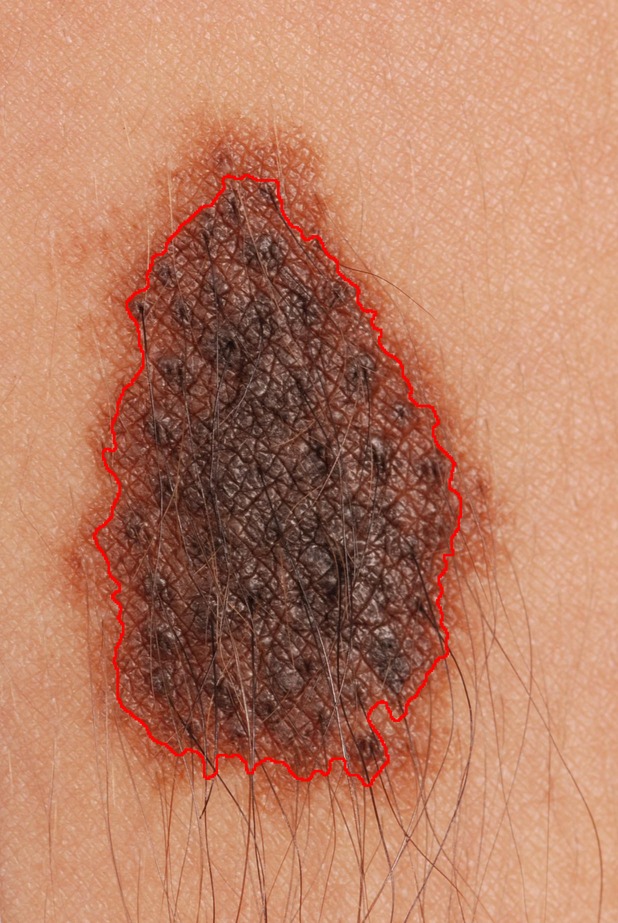}
}

\vspace{-1.6em}
\subfloat{
       \includegraphics[width=0.24\columnwidth]{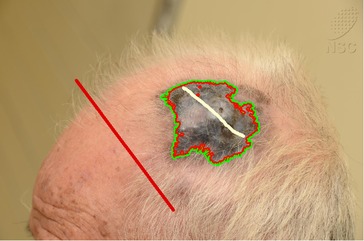}
}
\subfloat{
       \includegraphics[width=0.24\columnwidth]{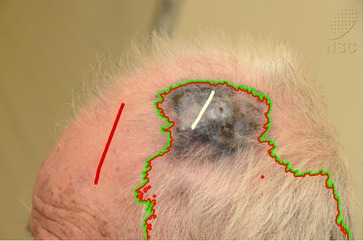}
}
\subfloat{
       \includegraphics[width=0.24\columnwidth]{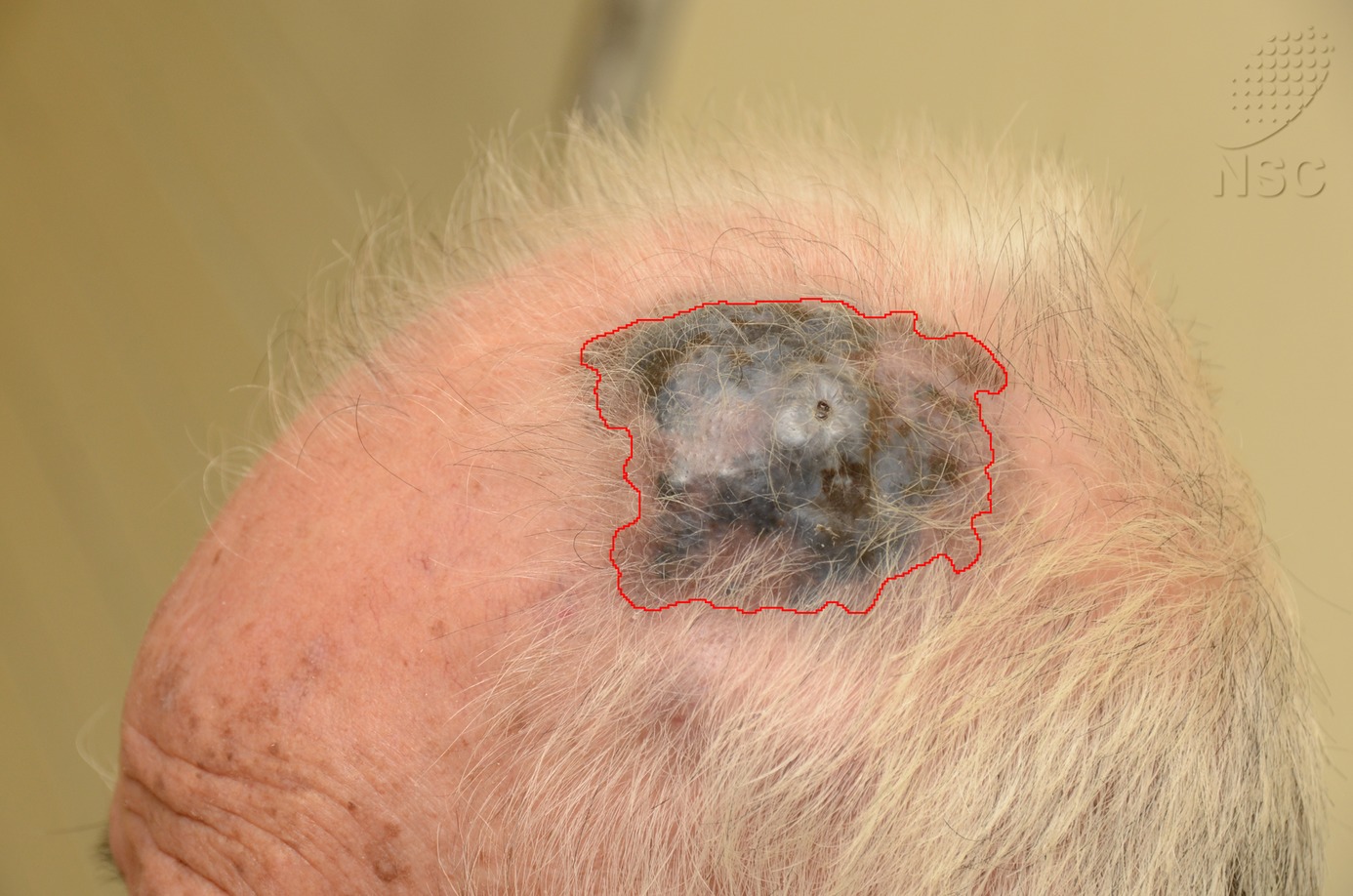}
}

\vspace{-0.7em}
\subfloat{
       \includegraphics[width=0.24\columnwidth]{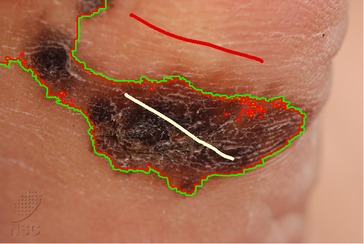}
}
\subfloat{
       \includegraphics[width=0.24\columnwidth]{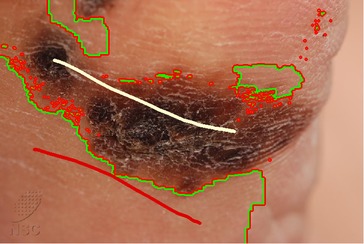}
}
\subfloat{
       \includegraphics[width=0.24\columnwidth]{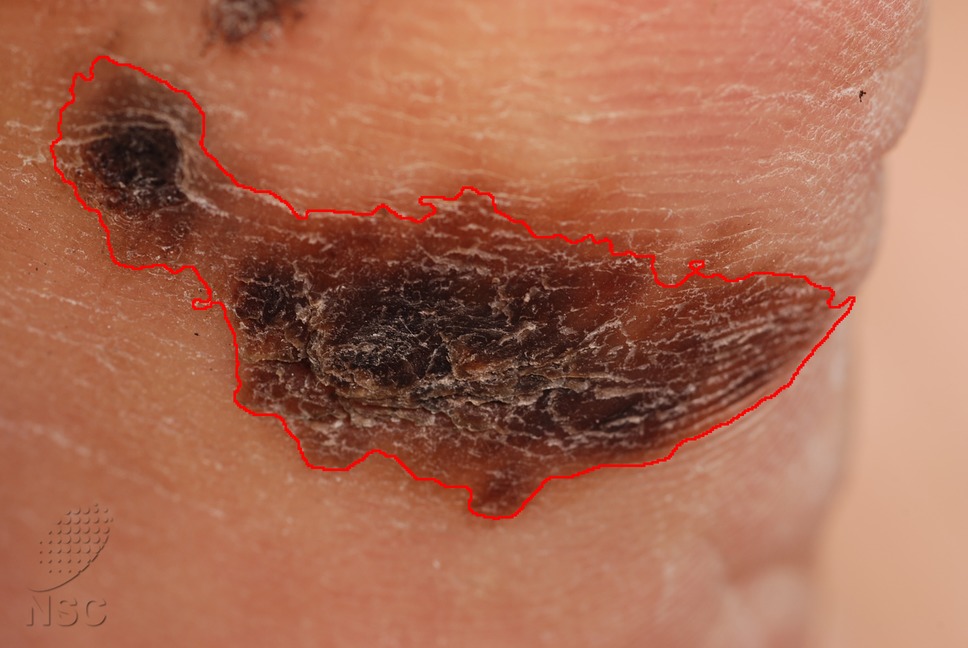}
}

\vspace{-0.7em}
\subfloat{
       \includegraphics[width=0.24\columnwidth]{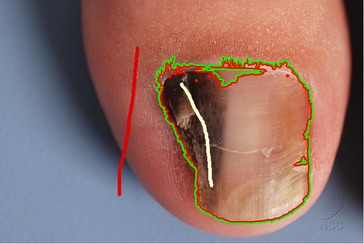}
}
\subfloat{
       \includegraphics[width=0.24\columnwidth]{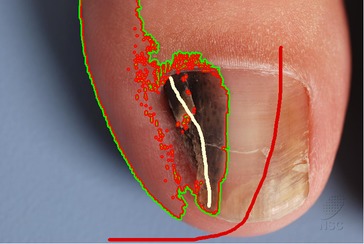}
}
\subfloat{
       \includegraphics[width=0.24\columnwidth]{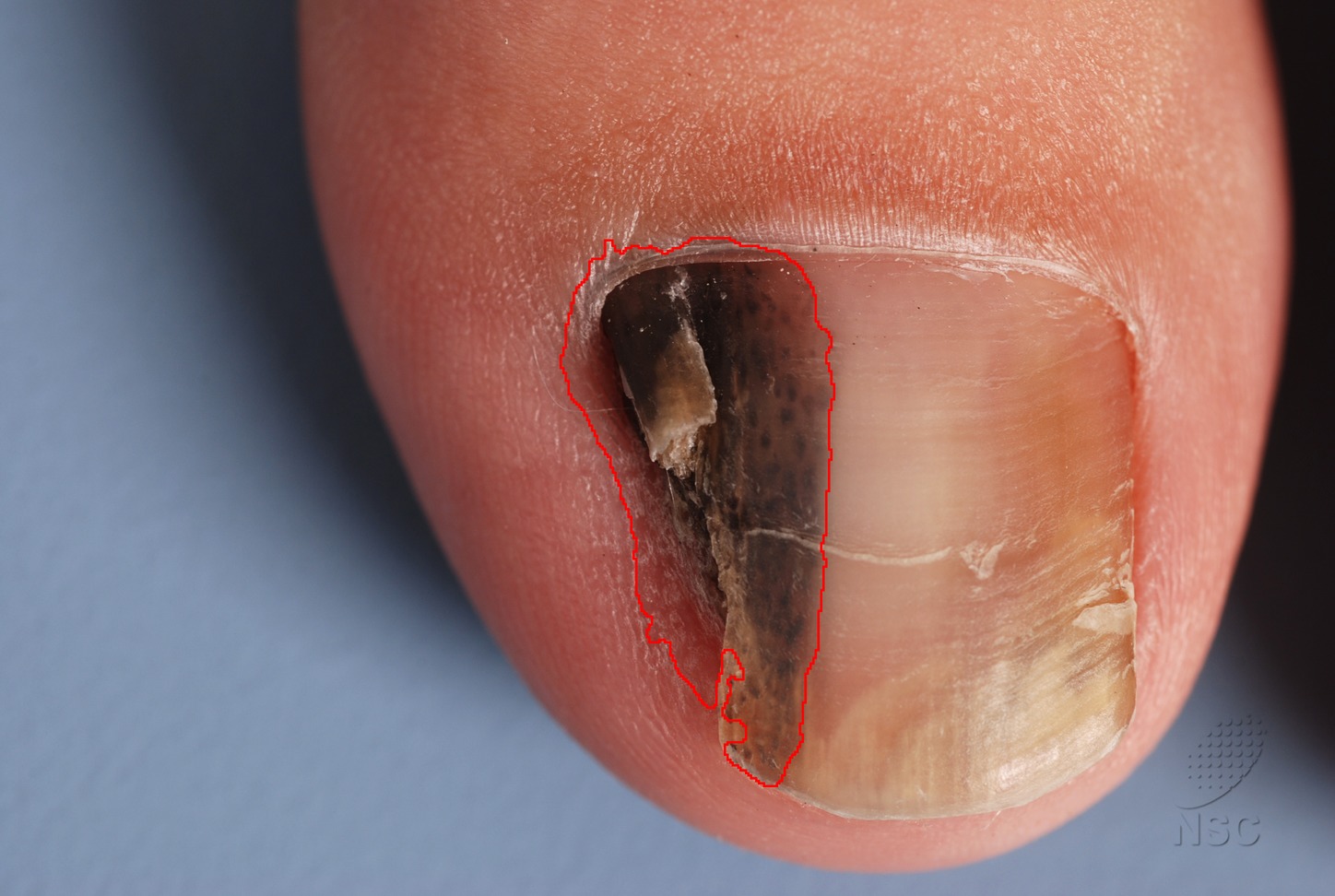}
}

\vspace{-0.7em}
\subfloat{
       \includegraphics[width=0.24\columnwidth]{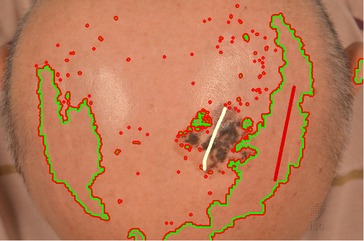}
}
\subfloat{
       \includegraphics[width=0.24\columnwidth]{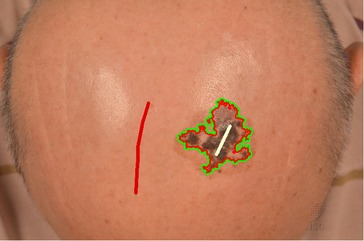}
}
\subfloat{
       \includegraphics[width=0.24\columnwidth]{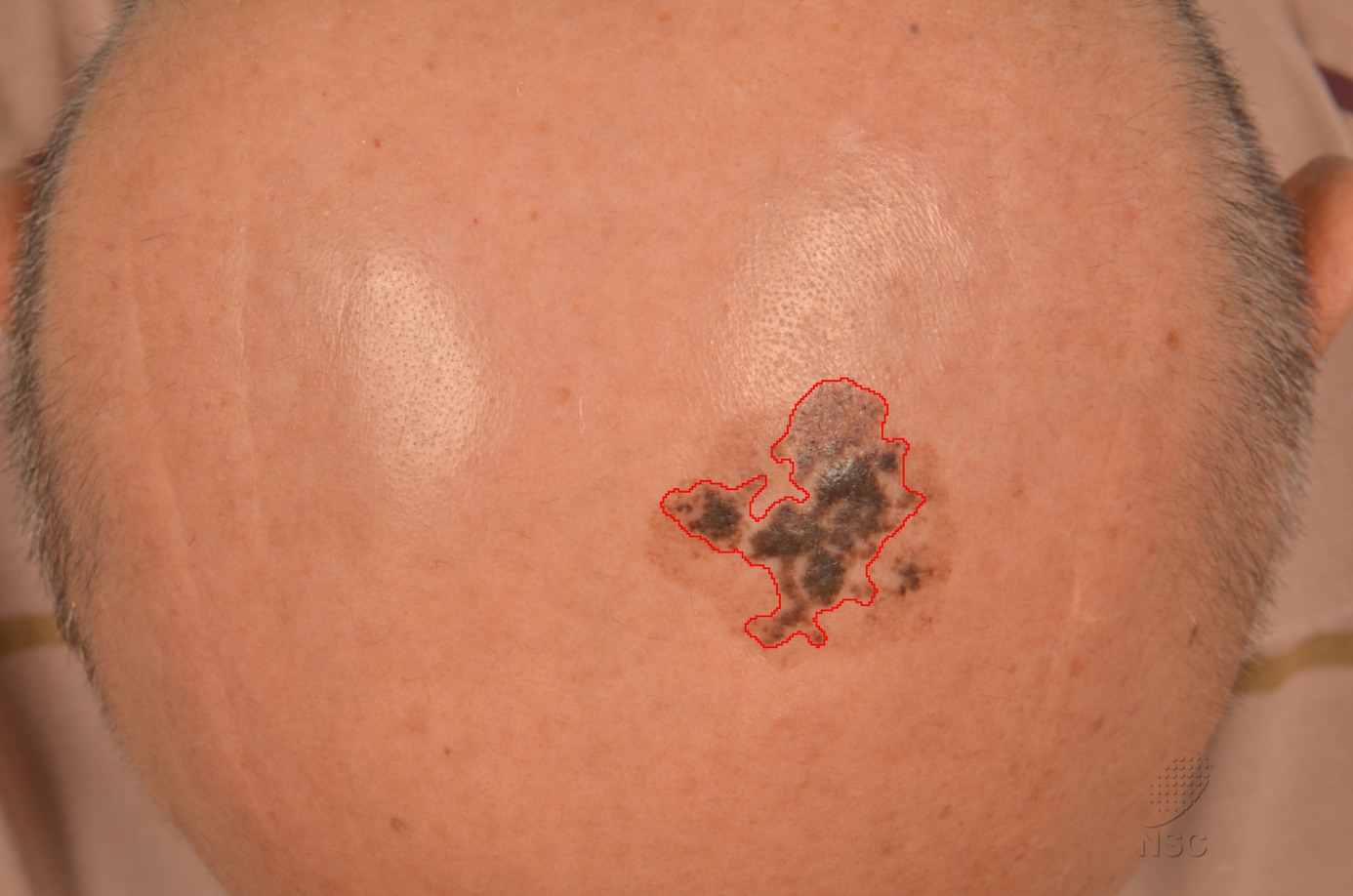}
}

\vspace{-0.7em}
\subfloat{
       \includegraphics[angle=90,origin=c,width=0.24\columnwidth]{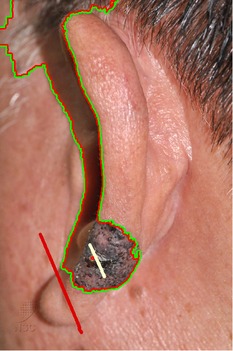}
}
\subfloat{
       \includegraphics[angle=90,origin=c,width=0.24\columnwidth]{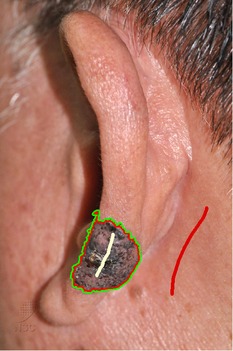}
}
\subfloat{
       \includegraphics[angle=90,origin=c,width=0.24\columnwidth]{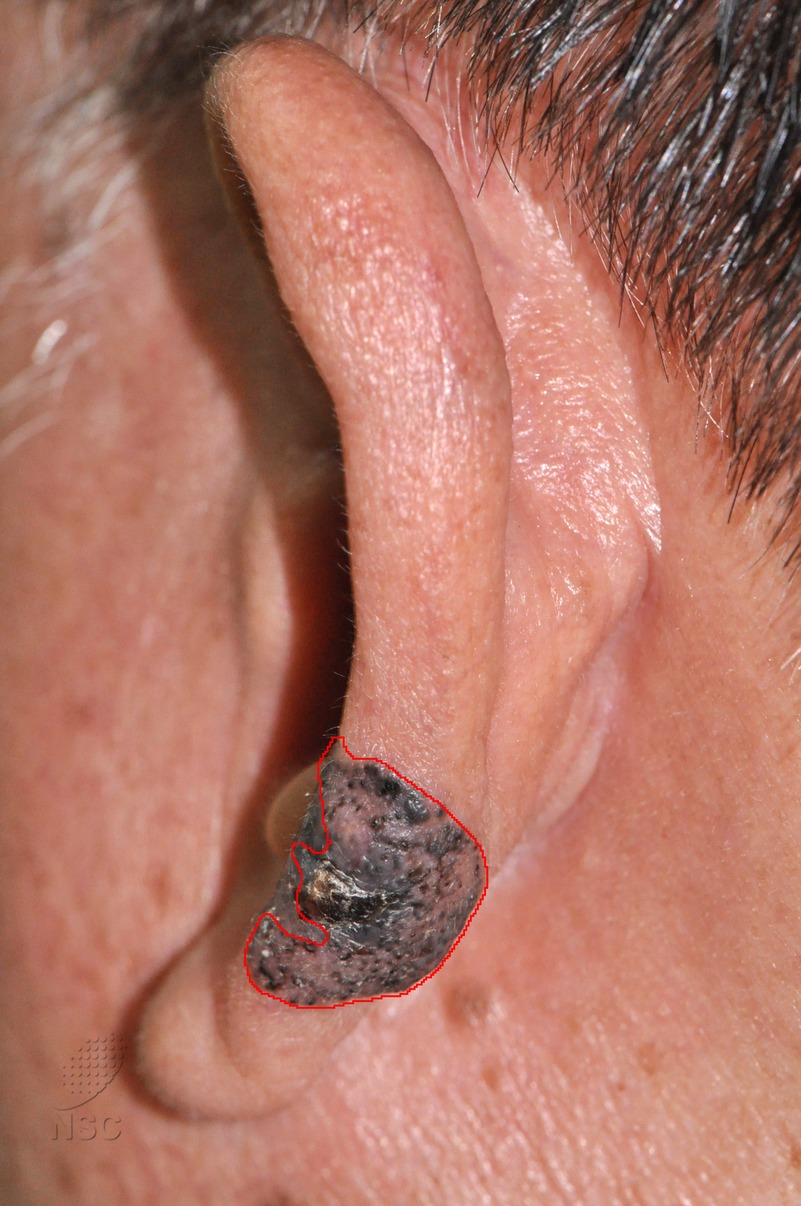}
}

\vspace{-1.6em}
\subfloat{
       \includegraphics[width=0.24\columnwidth]{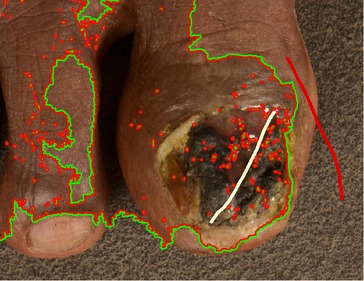}
}
\subfloat{
       \includegraphics[width=0.24\columnwidth]{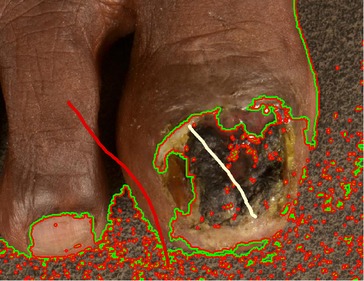}
}
\subfloat{
       \includegraphics[width=0.24\columnwidth]{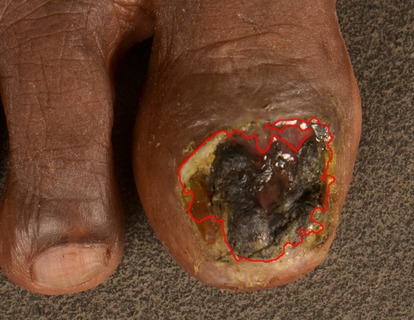}
}
\caption{\redd{Example of segmentation result using interactive method~\cite{gulshan2010geodesic}. Left \& middle columns: Interactive method with different initial strokes (red and white color). Right column: segmentation result using our proposed method.}}
\label{fig:graph-cut}

\end{figure}
\vspace{-1.5em}
}

\subsection{Feature Selection Results}
\label{subsec:fea-selec-result}

Our proposed feature selection tool is employed on individual feature categories, i.e., color (54 features), border (16 features) and texture (8 GLCM features and 1 edge density feature). Note that we have not applied the feature selection for the asymmetry category since it contains only one feature.  We also have not applied feature selection to LBP, as the LBP descriptors have high dimension and this requires a large dataset to estimate the probability distribution in (\ref{eq:MI_2}) to compute the mutual information.

 
In order to compute the mutual information, features should be first discretized. To discretize each feature, the original interval is divided into a number of equal bins. Feature values falling outside the interval are assigned to the extreme left or right bin. We run feature selection with the number of bins equal to $\left \{2, 3, 4, 5, 6 \right\}$ and let feature selection determine the optimal number of bins. The best classification accuracy is achieved when the number of bins is equal to 5.

The values of $n_i^o$ and $n_i^e$ in~(\ref{eq:ANM_Q}) are set to $50\%$ number of samples of class containing the sample $i$, and $\alpha$ in~(\ref{eq:hybrid}) is set to $0.4$.  For the CT feature the number of partitions considered in feature selection are $PA = \left \{4, 8, 12, 16\right\}$, while the number of subparts are $SP = \left \{2, 4, 8\right\}$. For the Border Fitting feature the number of lines analyzed are $nt=\left\{8, 12, 16, 20, 24, 28\right\}$.


Table~\ref{tab:res_fea_select} shows selected features in each category when MI-based criterion (\ref{eq:selection}) and our proposed criterion (\ref{eq:hybrid}) are used during the feature selection \cite{SUTD_EMBC2014}. The classification accuracy of different feature categories for MI-based criterion and our proposed criterion is given in Table~\ref{tab:FS_results}. 
\begin{table}[!t]
\small
 
\centering
\begin{tabular}{p{1.2cm}|p{3cm}|p{3cm}} 
\hline
\hline
Feature Category & Mutual Information        & Proposed\\ 
\hline
\multirow{4}{*}{color} & color triangle ($SP=16, PA=8$) ,  	& color triangle ($SP=16, PA=8$)  \\
& num\_red, num\_hue,			    &num\_gray, num\_hue\\
& num\_green                      & \\ 
\hline
\multirow{2}{*}{border} &mean of border                         & variances of border \\
& fitting ($nt=12$) & fitting ($nt=\left\{8, 12\right\}$)\\ 
\hline
\multirow{3}{*}{texture} &\multicolumn{2}{c} {edge density}\\ 
& \multicolumn{2}{c} {contrast of GLCM (64 quantization levels)}\\ 
& \multicolumn{2}{c}{correlation of GLCM (32 quantization levels)} \\
\hline
\hline
\end{tabular}
\vspace{-0.5em}
\caption{Selected features from each category when mutual information criterion and proposed criterion are used.}
\label{tab:res_fea_select}
\end{table}

\begin{table}[!t]
\small

\centering
\begin{tabular}{c|c c c|c c c} 
\hline
\hline
 {} &\multicolumn{3}{c|}{Mutual Information}     & \multicolumn{3}{c}{Proposed}\\  
\hline 
 {}        & color & border & texture & color & border & texture \\  
\hline
 Sens    & 94.00 & 82.55  & 84.55    & 94.18 & 79.27   & 84.55   \\
 Spec    & 86.00 & 66.00  & 76.67    & 90.00 & 76.00   & 76.67   \\
 Acc      & 90.00 & 74.27  & 80.61    & \textbf{92.09}  & \textbf{77.64}   & \textbf{80.61}   \\         
\hline                                                                                                                       
\hline
\end{tabular}
\vspace{-0.7em}
\caption{Feature selection results: the classification performance when MI-based criterion and our criterion are used to select features for individual categories. The classification results are with SVM classifier with 5-folds cross validation on SET2.  
}
\label{tab:FS_results}
\end{table}

{
From Table~\ref{tab:res_fea_select}, we can see that CT and Border Fitting features are always selected for both MI and proposed criteria. This confirms the efficiency of our proposed novel features.
}
For the texture category, both feature selection-based MI and the proposed method select same feature set, i.e., \textit{edge density, the contrast of 64 quantization level GLCM}, and \textit{correlation of 32 quantization level GLCM}. In following sections, we  refer these three selected features as \textit{GLCM+edge} features for distinguishing with LBP$_\text{S}$ features.

{
Additionally, Table \ref{tab:FS_results} shows that the proposed feature selection method can select more discriminative sets of features. In particular, for the color feature category, the proposed method can gain about 2\% accuracy improvement by selecting a set of only 3 features instead of a set of 4 features of MI-based method. 
For the border feature category, the MI-based criterion
achieve highest accuracy of 74.27\% when only one feature is
selected. By using MI-based criterion, we cannot obtain higher accuracy even when more border features are included.
On the other hand, the highest accuracy of the proposed criterion is 77.64\% when 2 border features are selected.
}

\begin{figure}[!t]
\centering
\subfloat[]{
       \includegraphics[width=3cm,height=3cm]{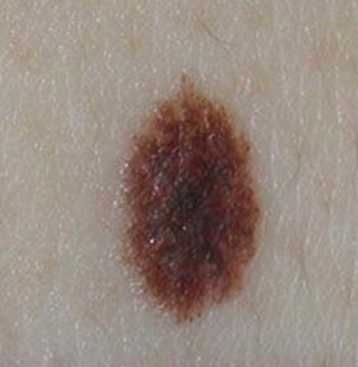}
}
\subfloat[]{
       \includegraphics[width=3cm,height=3cm]{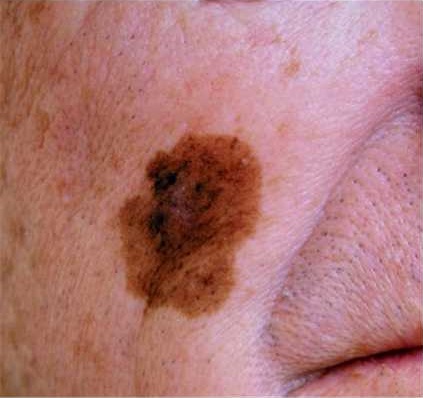}
}
\vspace{-1em}
\caption{Examples of (a) benign sample and (b) melanoma sample. Both have high color variation (i.e., high $num\_gray$) values.}
\label{fig:samples}

\end{figure}

{\bf The importance of using multiple features for each category:}    
We remark that we use multiple features for each feature category because of the large variation of the data (e.g., color variation). Using a single feature for each feature category is not enough to distinguish the melanoma and benign samples. For examples, although many melanoma samples usually have the variation in the color (e.g., the color spreads from white to dark), using only $num\_gray$ feature (the number of non-zero histogram bins on gray channel) does not give a high classification accuracy. It is  because there are some benign samples which also have the color spreading from white to dark. 
Two examples of the benign (left) and melanoma (right) samples are provided in Fig. \ref{fig:samples}.

Similarly, there are some benign samples which their color do not vary uniformly  from the center to the border. In those cases, using only Color Triangle feature is not sufficient for the classification. 
Hence, using both $num\_gray$ and Color Triangle features  helps overcome the data variation issue, and boost the classification accuracy.

We provide here the Table \ref{tab:color_classifier_results} which shows the classification accuracy, the sensitivity, and the specificity when the selected color features $(Color\ Triangle\ (SP = 16; PA = 8), num\_gray, num\_hue)$ are used individually or are used together. We can see that combining selected color features significantly boosts the accuracy. 

\begin{table}[!t]
\small
\centering
\begin{tabular}{{c c c c c}} 
\hline
\hline 
Feature  & $CT$ & $num\_gray$  & $num\_hue$ & Combined\\  
\hline 
Sens     & 85.21 & 76.92 & 88.39 & 94.18 \\
Spec     & 77.85 & 68.97 & 65.52 & 90.00 \\
Acc      & 81.53 & 72.94 & 76.96 & 92.09 \\  
\hline                                   
\hline
\end{tabular}

\caption{The classification performance on the SET2 for each selected color feature (the results are with SVM classifier with 5-folds cross validation). For the Color Triangle (CT) feature,  $SP = 16; PA = 8$.}
\label{tab:color_classifier_results}
\end{table}

We also provide here the Table \ref{tab:border_classifier_results} which shows the classification accuracy, the sensitivity, and the specificity when selected border features $(variances\ of\ Border\ Fitting\ for\ nt = 8,12)$   are used individually or are used together. 
The results show that combination of selected border features boosts the accuracy.  

\begin{table}[!t]
\small

\centering
\begin{tabular}{{c c c c}} 
\hline
\hline 
Feature & \multicolumn{2}{c}{Variances of Border} & Combined \\
        & $nt=8$ & $nt=12$  & \\  
\hline 
Sens   & 78.20  &68.96   & 79.27 \\
Spec   & 56.83  &57.69   & 76.00 \\
Acc    & 67.52  &63.32   & 77.64 \\  
\hline                                   
\hline
\end{tabular}
\vspace{-0.7em}
\caption{The classification performance on the SET2 for each selected border feature (the results are with SVM classifier with 5-folds cross validation).}
\label{tab:border_classifier_results}
\end{table}

\subsection{Classification Results}
 \label{subsec:classificationres}
\red{As discussed in} section \ref{subsec:classifier}, four binary SVMs are applied to the selected features from four feature categories (color, border, asymmetry, and GLCM+edge) and the $k$NN is applied for LBP$_\text{S}$ features. To aggregate useful information in individual classifiers, these classification results are processed further using several proposed \red{fusion methods}.

{\bf Performance of individual classifiers:}
For SVM classifiers, to estimate the generalization error of the models based on the selected features, we employ the 10-folds cross validation on the TEST SET, i.e., 154 segmented ROIs. SVMs return soft values which present the confidence level of a sample belongs to MM. By applying a threshold, we can obtain the hard output values, which is 0 (non-cancer, benign nevus) or 1 (cancer, MM). The columns ``Color'', ``Border'', ``Asymmetry'', ``GLCM+edge'' of Table~\ref{tab:classifier_results} show the performance of the system on TEST SET for each feature category at the threshold of 0.5.

\begin{figure*}[h]
\small
\centering
       \includegraphics[width=0.80\textwidth]{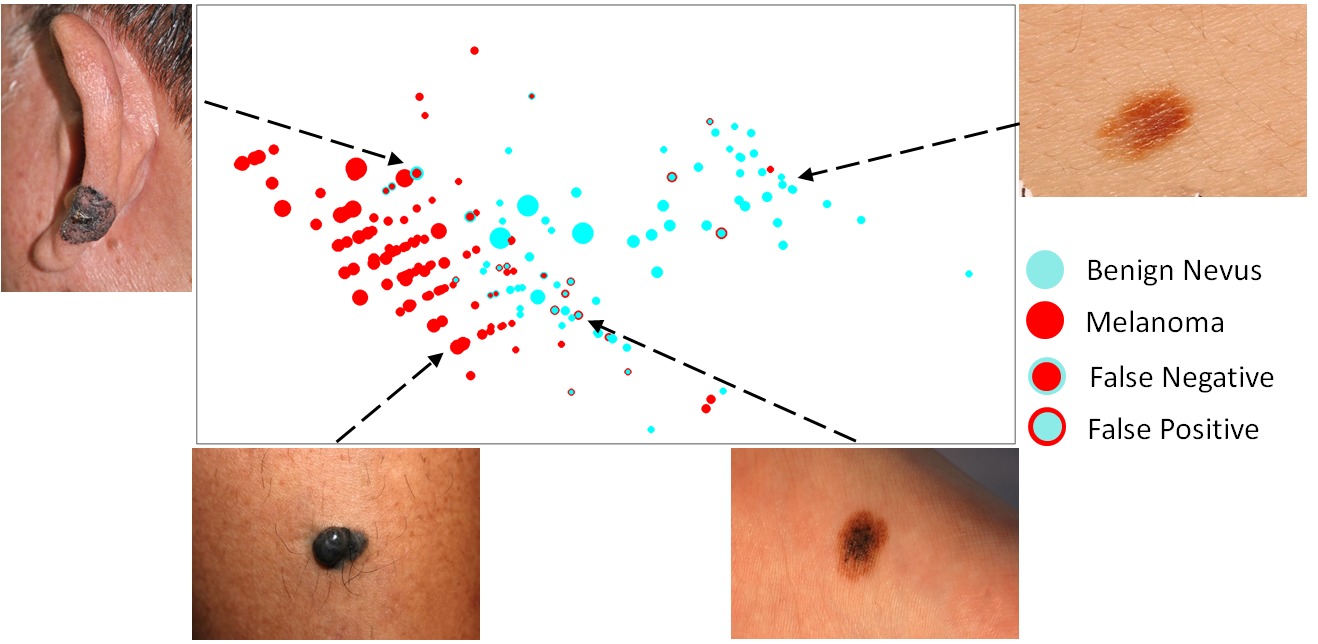}  
       \label{fig:fea_selec_border}
\caption{2D visualization of SVM output of the LCF after dimension reduction for TEST SET}
\label{fig:feat_viz}
\end{figure*}

Fig.~\ref{fig:feat_viz} shows the visualization of the SVM outputs of the LCF (color features) after dimension reduction and color-coding by the clinical labels.  The visualization reveals a good separation of regions corresponding to benign nevus and melanoma classes.  This suggests that distinct color characteristics are presences of different skin lesion classes which are captured by the selected features. Particular image samples in Fig.~\ref{fig:feat_viz} demonstrate how the proposed scheme allows a good separation of skin lesions.

As discussed, for the LBP$_\text{S}$ features, we use a $k$NN classifier.  We consider various settings, i.e., parameters for LBP: $R = \left\{1, 2\right\}, P = \left\{8, 16\right\}$; number of neighbors of $k$NN: $k=\left\{1, 2, 3\right\}$. We empirically achieve the best results for following the parameters: $R = 1$, $P =8$, and $k=2$. The $k$NN returns soft values that indicate the confidence level of a sample belonging to MM. Particularly with $k=2$, $k$NN returns 1 (or 0) when two nearest neighbors (NN) are MM (or benign); in the remaining situation, i.e., one NN is MM and another NN is benign, the distance from the test sample to its two NNs is utilized to calculate the soft output values. The classification results are showed in the ``LBP$_\text{S}$'' column of Table~\ref{tab:classifier_results} after applying a threshold of 0.5 on the soft output values.

{\bf Performance with fusion of individual classifiers:}
Furthermore, we employ three proposed fusion methods for fusing the results of the four classifiers, i.e., (Color, Border, Asymmetry, and GLCM+edge) or (Color, Border, Asymmetry, and LBP$_\text{S}$), as discussed in section \ref{subsec:classifier}. 

For the \red{fusion method} 2 (Weighed sum, 2a and 2b), we need to determine the weight values and the threshold to classify cancerous samples. Specifically, by running 3-folds cross validation on SET3, we achieve soft classification values. From these classification values, we calculate the weight, i.e., \textit{sensitivity@50\% specificity} and \textit{the Area Under Receive Operating Curve (AUC)}, using for the fusion. Additionally, the threshold is set to the value that maximizes the accuracy on SET3.
The weight values are useful in providing a better view of how feature categories contribute to the diagnosis process.

\begin{figure}[!t]
\begin{center}    
\includegraphics[width=0.8\columnwidth]{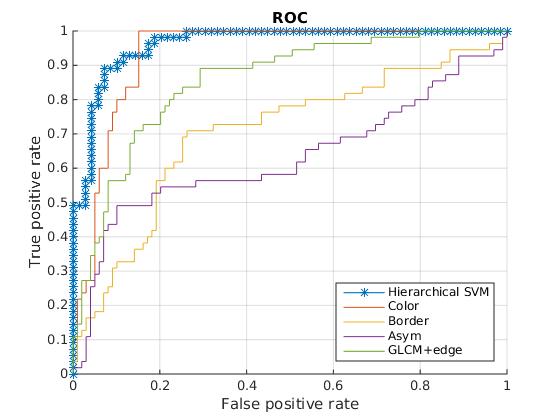}
\vspace{-1em}
\caption{Receiver Operating Curve (ROC) for four feature categories (color, border, asymmetry, and GLCM+edge) and hierarchical SVM \red{fusion method}. (Best view in color)}    
\label{fig:ROC}
\end{center}
\end{figure}

The results showed in Table~\ref{tab:fusing_results} and ROC in Fig. \ref{fig:ROC} demonstrate that fusion can improve the overall classification performance. We achieve the best \textit{Sens @ Spec >= 90 \%} of \textbf{89.09\%} for the \textbf{hierarchical SVM}  (color, border, asymmetry, and GLCM+edge) and \textbf{AUC weighted sum} (color, border, asymmetry, and LBP$_\text{S}$) methods. 
The results in Table~\ref{tab:fusing_results} suggest that different feature categories may not contribute equally in the diagnosis procedure, and proper combination and fusion of them significantly boost the accuracy.


\begin{table}[!t]
\small

\centering
\begin{tabular}{{c c c c c c}} 
\hline
\hline 
Feature   & Color & Border & Asymmetry & GLCM+edge & LBP$_\text{S}$ \\  
category  &    & & & & \\
\hline 
Length & 3      & 2      & 1     &  3       &  36     \\
Sens   & 96.36  & 69.09  & 47.28 & 69.09    & 79.10   \\
Spec   & 83.84  & 70.71  & 87.88 & 82.83    & 87.17   \\
Acc    & 90.01  & 69.90  & 67.58 & 75.96    & 83.14   \\   
\hline                                   
\hline
\end{tabular}
\vspace{-0.5em}
\caption{Individual Classifiers Performance: The classification performance on the TEST SET for each feature category, i.e., without classifier fusion. The length of each feature category is also provided.}
\label{tab:classifier_results}
\end{table}

\begin{table*}[!t]

\centering
\begin{tabular}{{c c c c c| c c c c}} 
\hline
\hline 
Feature categories  & \multicolumn{4}{c|}{(Color, Border, Asymmetry, and GLCM+edge)} & \multicolumn{4}{c}{(Color, Border, Asymmetry, and LBP$_\text{S}$)} \\  
\hline 
\red{Fusion methods} & 1 & 2a & 2b & 3 &  1 & 2a & 2b & 3 \\
\hline 
Sens  & 94.55 & 94.55  & 94.55 & 92.73 & 100.00 & 100.00 & 89.09 & 98.18  \\
Spec  & 83.84 & 74.75  & 83.84 & 85.51 & 84.85  & 84.85  & 86.87 & 82.83  \\
Acc   & 89.19 & 84.65  & 89.19 & 89.12 & 92.42  & 92.42  & 87.98 & 90.51 
\\ \hline
Sens@ Spec >= 90\% & 72.73 & 76.73 & 87.27 & \textbf{89.09} & 67.27 & 73.72 & \textbf{89.09} & 80.00 \\
\hline                                   
\hline
\end{tabular}
\vspace{-0.7em}
\caption{The performance of the proposed system with fusion of individual classifier results, using different proposed fusion methods: 1, 2a, 2b, and 3 are \textit{Non-weighted sum}, \textit{Sens@ 50\%spec weighted sum}, \textit{AUC weighted sum} and \textit{hierarchical SVM} respectively.} 
\label{tab:fusing_results}
\end{table*}

\redd{\subsection{Comparison with existing methods}
 \label{subsec:comparison}
\begin{table}[!t]

\centering
\begin{tabular}{{c c c| c c| c c}} 
\hline
\hline 
   & \multicolumn{2}{c|}{Color} & \multicolumn{2}{c|}{Texture} & \multicolumn{2}{c}{Color+Texture}  \\  
 &\cite{mednode} &Ours  &\cite{mednode} &Ours &\cite{compare_ICIP16} &Ours\\
\hline 
Sens &74 &81 &62 &66  &82 &84 \\
Spec &72 &73 &85 &85  &71 &72 \\
PPV  &64 &66 &74 &75  &67 &70 \\	
NPV  &81 &85 &77 &79  &85 &87 \\
TAcc &73 &75 &76 &78  &76 &77 \\   
\hline                                   
\hline
\end{tabular}
\vspace{-0.7em}
\caption{\redd{The comparative results (in \%) between MED-NODE \cite{mednode} and \cite{compare_ICIP16} and our proposed system. MED-NODE \cite{mednode} reports results for color and texture features independently. \cite{compare_ICIP16} reports the results when color and texture features are combined together. The results of MED-NODE \cite{mednode} and \cite{compare_ICIP16} are cited from the corresponding papers. We also follow the experimental setups of the corresponding papers.}}
\label{tab:comparison}
\end{table}
}

In addition, we compare our proposed melanoma detection system  with some recent melanoma detection methods: MED-NODE \cite{mednode} and  \cite{compare_ICIP16},
on the publicly available UMCG dataset from Department of Dermatology of University Medical Center Groningen (UMCG) \cite{UMCG}. 
UMCG is a challenging dataset.
The dataset consists of 70 melanoma and 100 benign images. MED-NODE \cite{mednode} performs the training on 45 randomly selected images (20 melanoma and 25 benign images, respectively) and reports the results on the remaining 125 images. \cite{compare_ICIP16} performs the training on 125 randomly selected images (50 melanoma and 75 benign images, respectively) and reports the results on the remaining 45 images. 
For fair comparison, we apply exactly the same data division when comparing to the corresponding methods. It is worth noting that both MED-NODE \cite{mednode} and \cite{compare_ICIP16} use color and texture features to describe the image. MED-NODE \cite{mednode} reports the classification results for each feature category (i.e., color and texture) independently, while \cite{compare_ICIP16} reports the classification results when both color and texture features are combined together. 
For fair comparison, we also use only our proposed color and texture features in this experiment.

Our system first performs the segmentation (Section III.A), the feature computation for segmented lesion (Section III.B), and the feature selection (Section III.C) on the training dataset. We then report the results with the proposed hierarchical SVM classifier (Section III.D). 
Similar to \cite{mednode} and \cite{compare_ICIP16}, we report following metrics: Sensitivity (Sens), Specificity (Spec), Positive Predictive Value (PPV), Negative Predictive Values (NPV), and Total Accuracy (TAcc - which is the number of correct classified samples over the number of testing samples) to assess the performance of our system. 
Our results are the average of 50 random splits of training/testing sets.  The experimental results presented in Table \ref{tab:comparison} show that our proposed method outperforms recent methods \cite{mednode} and \cite{compare_ICIP16}. It is worth noting that our proposed method is designed to work on mobile device, while it is not clear if MED-NODE \cite{mednode} and \cite{compare_ICIP16} can work on a resource-constrained mobile device. 

Furthermore, we would like to remark that: 
1) While there are other mobile applications (Apps) for melanoma detection using smartphones (e.g., SkinVision \cite{skinvision}), we found that they can only take an input image from the phone's camera directly.  In particular, these Apps do not allow a user to load a pre-captured image for testing. Therefore, it is not possible to compare their accuracy using a common dataset.
2) Most of these Apps have not been thoroughly tested and their performance is unknown. For a few Apps which have been tested, their systems were evaluated using their own proprietary images \cite{Maier:2015}, making fair comparison with our work difficult. Therefore, we focus on comparison with recent works \cite{mednode,compare_ICIP16} that have been evaluated using the common dataset, i.e., UMCG  \cite{UMCG}.
\\
\redr{
\textbf{Computational complexity analysis and comparison with recent methods:} Several complete systems have been recently proposed for non-dermatoscopic image-based melanoma detection:~\cite{mednode,compare_ICIP16}, and the proposed one in this work. Although these systems have similar general stages, i.e., lesion segmentation, feature extraction, and classification, these systems use very different techniques at different stages. 


For the lesion segmentation, we use MST and Otsu methods, while  \cite{mednode,compare_ICIP16} used K-means.
Given a lesion segmentation, our method uses hierarchical SVM to classify the lesion by using $9$ selected features (3 color features, 2 border features, 1 asymmetry feature, and 3 texture features). On the other hand,  \cite{compare_ICIP16} uses Neural Network (NN) for classification by using 10 color features.
 \cite{mednode} combines two different classifiers.  The authors use the Cluster-based Adaptive Metric (CLAM) \cite{CLAM} to classify 12 color features. Additionally, the authors also use Unbiased Color Image Analysis Learning Vector Quantization (Unbiased CIA-LVQ) \cite{CIA-LVQ} method to classify texture features.
 
%
%
Here we perform a complexity analysis and comparison at different stages of the processing:

\textbf{Lesion segmentation stage: } Let us define the number of pixels in the skin region as $N$. 
For the lesion segmentation, we aim to segment the skin pixels into 2 clusters, i.e., lesion or non-lesion. We use MST and Otsu, and both methods have $\mathcal{O}(N)$ complexity. The works \cite{mednode,compare_ICIP16} use K-means (in  color space) to segment the skin region into two clusters. Hence the complexity of~\cite{mednode,compare_ICIP16} for lesion segmentation step is $\mathcal{O}(NI)$ in which $I$ is the number of iterations of K-means\footnote{The general complexity of K-means is  $\mathcal{O}(NICd)$,  in which $N$ is the number of data points (i.e., pixels); $I$ is the number of iterations of K-means; $C$ is the number of clusters; $d$ is dimension of each data point. Here $C$ and $d$ are very small in comparison to $N$, i.e., $C=2$ and $d=3$, respectively.}. Because $I$ is usually very small in comparison to $N$, hence at this stage, \cite{mednode,compare_ICIP16} and ours have comparable complexity.

\textbf{Feature extraction stage: } Let us define the number of pixels in the lesion region as $N$. For the feature extraction, among our $9$ features, Color Triangle (CT) requires the most computation. Specifically, its  complexity is $\max (\O(PA^2 \times SP),\O(SP \times N))$, in which $PA$ and $SP$ are the number of parts and subparts of the CT feature. In our experiments, by using $PA=8$ and $SP=16$ (Table~\ref{tab:res_fea_select}) which are much less than the number of pixels in the lesion, hence the complexity of the CT feature is  $\O(SP \times N)$. 

In \cite{compare_ICIP16}, the most complex feature is the Color Variation which is computed by Fuzzy C-means (FCM) whose complexity is $\mathcal{O}(NI)$, in which $I$ is the number of iterations of FCM\footnote{The general complexity of FCM is  $\mathcal{O}(NIC^2d)$, in which $N$ is the number of data points (i.e., pixels); $I$ is the number of iterations of FCM; $C$ is the number of clusters; $d$ is dimension of each data point. Here $C=2$ and $d=3$, respectively.}.

In \cite{mednode}, for color features, it uses means and standard deviations of color channels, which have the complexity of $\mathcal{O}(N)$. For texture feature extraction, \cite{mednode} processes as follows: (i) the salient point detector in \cite{WALTHER20061395} is used to detect the 50 most interesting positions (pixels); (ii) at these positions, 15 $\times$ 15 patches are drawn and are used as texture features. 
Regarding the salient point detection, it is a multi-stage processing, i.e., firstly,  Gaussian filters are applied to input at multi-scales to produce multi-scale feature maps. The feature maps are then combined together. Finally, the combined feature map is classified by a neural network to produce salient locations. By using a multi-stage processing which consists of several complex components, i.e., multi-scale filtering, classification-based neural network, the computational cost of this detector is high, so is the whole process.
In summary, for the feature extraction stage, the complexities of our method and \cite{compare_ICIP16} are comparable and less than the complexity of \cite{mednode}.

\textbf{Classification stage:} For the classification, our system uses hierarchical SVM comprised of two stages of classifiers. In the first stage, we use 4 RBF-kernel SVM (one for each feature category). In the second (last) stage, another RBF-kernel SVM is used to fuse the soft decisions of the first stage's classifiers. The complexity of a RBF-kernel SVM is $\mathcal{O}(sd)$, where $s$ is the number of support vectors, which is around 50 in our experiments, and $d$ is the number of features.
In particular, in the first stage, we use $d=\{3, 2, 1,3\}$ for color, border, asymmetry, and texture features respectively, and $d=4$ for the second stage's classifier. 

The work \cite{compare_ICIP16} uses 2 layer feedforward neural network with the sigmoid activation. Hence, its  complexity is $\mathcal{O}(dH)$, in which $d=10$ is the number of features and $H=15$ is number of nodes in a hidden layer. 

In \cite{mednode}, the complexities of CLAM and Unbiased CIA-LVQ classifiers are 
$\mathcal{O}(d^2)$, where $d$ is the numbers of features. 
The input of CLAM is only 12 color features, so it is expected to have a fast inference. However, for Unbiased CIA-LVQ classifier, it first performs the classification for each of 50 patches in which the patch size (feature size) is $d = 15\times15$ pixels.  It then fuses these 50 outputs to produce final classification result. Hence, the classification stage of \cite{mednode} is dominated by Unbiased CIA-LVQ classifier which is higher complexity than the classifiers of \cite{compare_ICIP16} and ours. 



Overall, according to the complexity analysis of the main stages, our method's complexity is comparable to~\cite{compare_ICIP16} and is smaller than that of \cite{mednode}. Furthemore, it is worth noting that the processing time of \cite{mednode,compare_ICIP16} on resource-constrained smartphones has not been examined (the source codes of \cite{mednode,compare_ICIP16} are not available). On the other hand, for our method, the running time on a smartphone is presented in Section \ref{subsec_HCIres}. Specifically, on the device ``Samsung Galaxy S4 Zoom smartphone, with Dual-core CPU running at 1.5GHz Cortex-A9, GPU: Mali-400, RAM: 1.5GB, and storage memory of 8GB'', our method takes less than 5 seconds to process an image. 
}

\subsection{Mobile Implementation and Human-Computer Interface (HCI) Design}
\label{subsec_HCIres}
We implement our proposed image analysis engine and the entire system on a consumer electronic mobile device: Samsung Galaxy S4 Zoom smartphone, with Dual-core CPU running at 1.5GHz Cortex-A9, GPU: Mali-400, RAM: 1.5GB and storage memory of 8GB. The features of the backside camera (the one used during the tests) are: 16 MP, image size: $4608\times3456$ pixels, with autofocus and $10\times$ optical zoom. 
We measured the average processing time spent for each image and it is less than 5 seconds. It is worth pointing out that, the mobile phone implementation of the algorithm has not been explicitly parallelized using the available GPU. 

{\bf HCI Design Study.} We conducted a study with an aim to understand the best HCI design  principles for this applications.  The subjects selected for the study were recruited through a local community center (i.e., older  participants with $\mu$=63.0 and $\sigma$=4.76) and from a local  university (i.e., young  participants  with $\mu$=24.1 and $\sigma$=3.22).  The  whole  evaluation study lasts for 1.5 hours. 




When entered the specific test, each participant was required to take a photo of his/her arm skin and then to click the ``Start Diagnosis'' button. After that, a progress bar was shown to indicate the processing time.
The participant was told that he/she could stop the diagnosis process by clicking the ``Give up to know the result'' button at any point during the processing time.  After the session, the researcher debriefed to the participant that the participant should consult professional doctors if he/she is interested in understanding his/her health condition towards the presented disease. 

After that, we conducted semi-structured interviews with participants exposed to our prototype. Since our application (a self-diagnosis application) can be considered as a subset of personal informatics systems, we constructed our interview questions based on the structure of stage-based model proposed in \cite{forlizzi:chi:2010}.  We aimed to identify the design challenges across the entire adoption process. 
Some important aspects regarding acceptance, collection, integration, reflection, and action, which have emerged from the study are:
\begin{itemize}
\item \textit{Acceptance}. One HCI issue that has emerged is the ``perceived harmfulness'' of the system.
Despite our application being implemented on a smartphone, older participants raised questions and concerns about the potential harmfulness of the application. Older adults may internally associate our application with the hospital used diagnostic machines,  and they questioned about whether some harmful detection methods (e.g., X-Ray, radiation) were used.  Even for the older adults who have been using smartphones for a long period they have similar concerns.
\item \textit{Reflection}. Some participants expressed  concerns regarding the presentation of the diagnosis results, as P23 mentioned: ``I want to know the result, but I don't want the application directly tells me the conclusion. Instead, could it show me the percentage of risk ... or how many people have been diagnosed the disease by this application and the possibility of being proved fine finally?''  P5 mentioned: ``It is very scary if the machine tells me I have skin cancer ...''.
\item \textit{Action}. For an effective self-diagnosis application, it is important to improve the trustfulness of the results and \red{encourage users} to make certain action (e.g., visit a hospital for a formal check). Some participants expressed the expectation towards the format of the diagnosis result, as P25 mentioned:  ``I hope the application can provide me more information, such as how the machine gets the diagnosis result. Which features, or what cause the disease happened...otherwise, I couldn't make decisions about the next step.''
\end{itemize} 
These HCI issues need to be addressed for successful adoption of the proposed system.

\section{Discussion}
\label{sec:dis}

The prototype presented in this paper has several limitations.  First, it is necessary to further validate the system with a large database.
It is worth pointing out that currently there are no publicly available clinical large datasets of camera phone images.  It is very challenging and time consuming to generate large datasets (especially for the MM cases), as these images need to be analyzed histopathologically to produce the gold standard.

Second, on some occasions, the images acquired by the mobile devices can be heavily affected by distortions, in particular, motion blur and \red{illumination variations} \cite{Distortions1, Distortions2}. These distortions can change the skin lesion appearance and smooth the border of the skin mole, confusing the segmentation and classification algorithms that our current system cannot compensate them.  It is desirable to include a module to detect such distorted images and alert users in such situations.

Third, the feature categories exploited by our system are extracted merely from the skin images and attempt to incorporate the main dermatologic signs. 
Our system can be enhanced by incorporating  
patient-derived clinical data (e.g., age, gender, size and location of the lesion).
We believe that this side information will improve the sensitivity and specificity of the proposed system.

\vspace{-0.7em}
\section{Conclusion}
\label{sec:con}

We propose an accessible mobile health-care solution for melanoma detection, using mobile image analysis.
The main characteristics of the proposed system are: an efficient hierarchical segmentation scheme suitable for the resource-constrained platform, a new set of features which efficiently capture the color variation and border irregularity from the smartphone-captured image, and a new mechanism  for selecting a compact set of the most discriminative features. The experimental results based on 184 camera images demonstrate the efficiency of the prototype in accurate segmentation and classification of the skin lesion in camera images. We foresee several possible usage scenarios for the current solution: it could be employed by the general public for preliminary self-screening or it can assist the general physicians during the diagnosis process.

In addition to the technical development, we paid attention also to understand the usability and acceptance challenges.
For this purpose, we have investigated the HCI design issues through an exploratory case study and semi-structured interviews. 
Our study discovered several important HCI issues that should be addressed in future work.



\vspace{-0.7em}
\section*{Acknowledgment}
{\footnotesize{
The authors would like to thank Dr. Martin Chio and Dr. Michelle Liang from NSC for providing us clinical opinions and the dataset, and helping to validate the system. Furthermore, we would like to thank Prof. Susan Swetter, from Stanford University Medical Center and Cancer Institute, for her valuable comments during this research.
The material reported in this article is supported in part by the MIT-SUTD International Design Centre (IDC). Any findings, conclusions, recommendations, or opinions expressed are ours and do not necessary reflect the views of the IDC.}}

\bibliographystyle{IEEEtran}
\bibliography{ref2-bk}

\end{document}